\documentclass[twocolumn]{article}

\usepackage{style/PRIMEarxiv}
\usepackage[utf8]{inputenc} 
\usepackage[T1]{fontenc}    
\usepackage{hyperref}       
\usepackage{url}            
\usepackage{booktabs}       
\usepackage{amsfonts}       
\usepackage{amsmath}
\usepackage{nicefrac}       
\usepackage{microtype}      
\usepackage{lipsum}
\usepackage{fancyhdr}       
\usepackage{graphicx}       
\usepackage{adjustbox}
\usepackage[most]{tcolorbox}
\graphicspath{{figures/}}
\usepackage{style/custom_listings}
\usepackage{cleveref}
\usepackage[inline]{enumitem}
\usepackage[table,xcdraw]{xcolor}
\usepackage{multirow}

\usepackage{style/references}
\setlist{leftmargin=*, itemsep=0.3em, topsep=0.3em}
\definecolor{highlightorange}{HTML}{FFCC99}

\newcommand{\databar}[5]{
  \begin{tikzpicture}
    \fill[#4!40] (0,0) rectangle ({#1/#2*#3},0.5);
    \draw (0,0) rectangle (#3,0.5);
    \node at (#3/2,0.25) {\small #5};
  \end{tikzpicture}
}

\definecolor{pastel1}{HTML}{ADD8E6}
\definecolor{pastel2}{HTML}{F2A65A}
\definecolor{pastel3}{HTML}{6F8F72}

\graphicspath{{figures/}}
\definecolor{highlightorange}{HTML}{FFCC99}

\newcommand{\repolink}{\href{https://github.com/pablomiralles22/OSST}{pablomiralles22/OSST}}

\title{One-shot Style Transfer LLM log-probabilities for Authorship Attribution and Verification}

\author{
  \textbf{Pablo Miralles-González\textsuperscript{1}},
  \textbf{Javier Huertas-Tato\textsuperscript{1}},
  \textbf{Alejandro Martín\textsuperscript{1}},
  \textbf{David Camacho\textsuperscript{2}}
\\
\\
  \textsuperscript{1}Department of Computer Systems,
  Technical University of Madrid,
  Madrid, Spain
\\
  \textsuperscript{2}United Arab Emirates University,
  Al Ain, United Arab Emirates
\\
}

\begin{document}
\twocolumn[\maketitle]

\begin{abstract}
Computational stylometry studies writing style through quantitative textual patterns, enabling applications such as authorship attribution, identity linking, and plagiarism detection. Despite the relevance of language modeling to these tasks, the pre-training of modern large language models (LLMs) has been underutilized in authorship attribution and verification. We introduce an unsupervised framework that uses the log-probabilities of an LLM to measure style transferability between two texts. This framework takes advantage of the extensive Causal Language Modeling (CLM) pre-training, one-shot capabilities and scale of LLMs, avoiding explicit supervision. Our methods substantially outperform prompting-based unsupervised baselines in authorship verification at similar model sizes, and is competitive with or improves contrastive baselines in most settings with sufficient model scale. We further observe strong performance across non-English languages. The effectiveness of the proposed framework improves consistently with increasing model scale. In the case of authorship verification, we propose an additional mechanism that increases test-time computation to improve accuracy; enabling flexible trade-offs between computational cost and task performance.
\end{abstract}


\section{Introduction}\label{sec:intro}

Computational stylometry is the quantitative and computational analysis of writing style, using patterns in vocabulary, syntax, punctuation, and character sequences to identify, verify, or cluster authors based on textual data~\cite{daelemans2013ExplanationComputational}.
Authorship analysis has been widely applied across diverse domains, with applications spanning both forensic and humanities contexts. In forensics, stylistic and linguistic patterns have been used to analyze terrorist discourse \cite{abbasi2005ApplyingAuthorship}, link user identities across online platforms \cite{shu2017UserIdentity}, detect plagiarism \cite{stamatatos2011Plagiarismauthorship}, and identify coordinated disinformation campaigns \cite{stiff2022Detectingcomputergenerated}. Authorship methods also play a key role in the humanities, supporting tasks such as the attribution of pseudepigraphic texts \cite{koppel2013Automaticallyidentifying} and the analysis of anonymous literary works \cite{stover2016Computationalauthorship}.

Authorship analysis is a long-standing challenge in NLP. Early approaches used hand-crafted stylometric features~\cite{stamatatosAuthorshipVerificationReview2016,lagutinaSurveyStylometricText2019}. But the advent of pre-trained transformers~\cite{vaswani2017AttentionAll,devlin2019BERTPretraining} led to the use of supervised encoder models~\cite{fabien-etal-2020-bertaa,zangerleOverviewMultiauthorWriting2023}.
We also find work fine-tuning individual language models for candidate authors~\cite{huangAttributingAuthorshipPerplexity2025,barlas2020CrossDomainAuthorship}, measuring the probability assigned to texts by the language model. While this approach tries to take advantage of the pre-training of LLMs, it is expensive to apply, and not useful in low resource contexts like authorship verification tasks.

Recent work has increasingly focused on learning author embeddings (see, e.g.,~\cite{rivera-soto2021LearningUniversal,sawatpholTopicRegularizedAuthorshipRepresentation2022,patelStyleDistanceStrongerContentIndependent2025}), mapping texts to dense vector representations that encode stylistic features. These approaches also train on labeled data, but optimize contrastive objectives. These techniques yield representations that are interpretable and transferable to many authorship tasks, including author profiling.

In this work we focus on non-supervised approaches. The emergence of large language models (LLMs) has led to prompt-based unsupervised stylistic analysis~\cite{hung2023WhoWrote,huang2024CanLarge}, offering an alternative without supervision but facing poor performance (at least at moderate model sizes; see~\Cref{tab:results-verification}) and challenges in attribution due to context length constraints.

\begin{figure*}[!ht]
    \centering
    \includegraphics[width=0.95\linewidth]{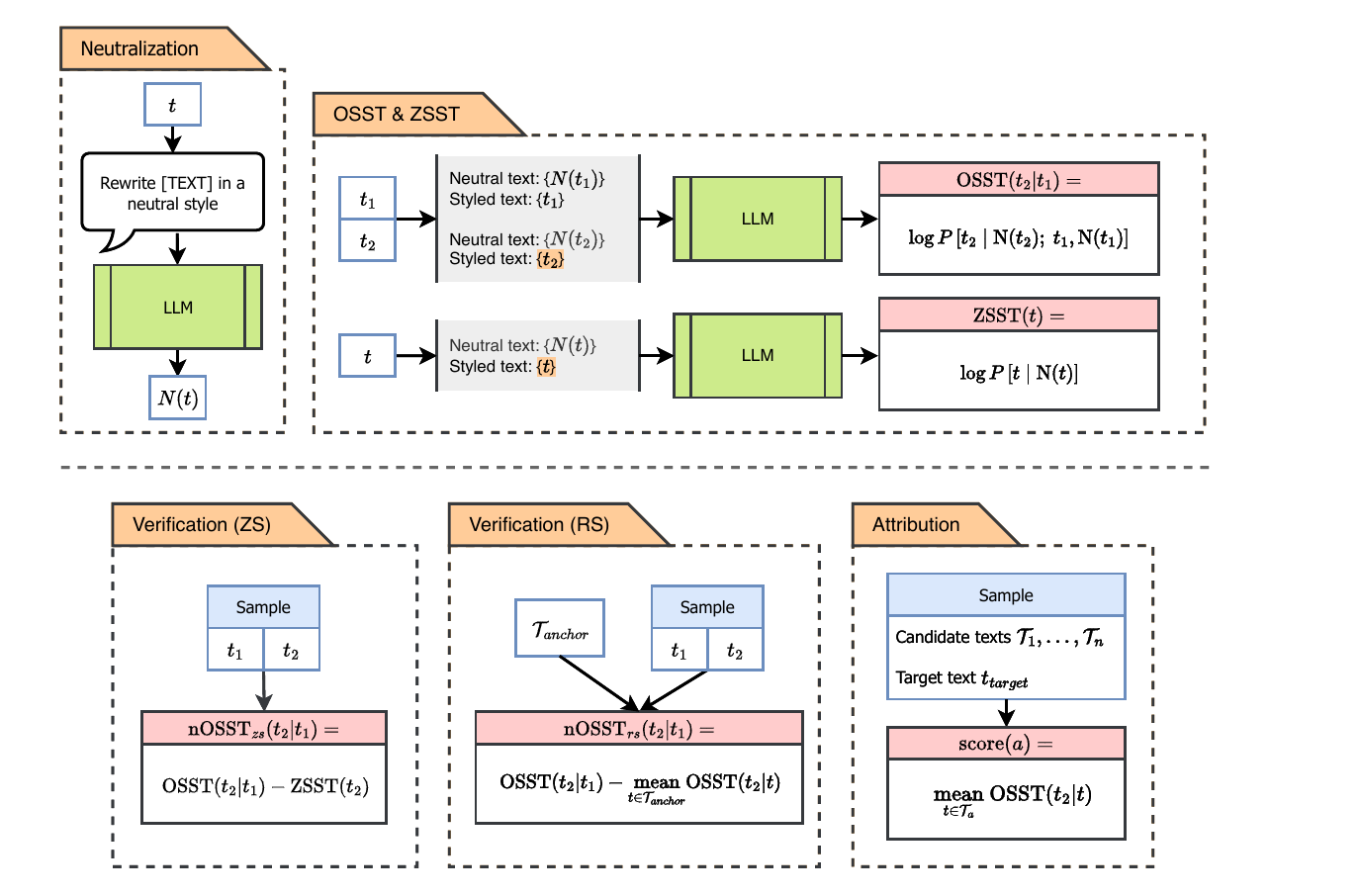}
    \setlength{\fboxsep}{0pt} 
    \caption{High-level overview of OSST-based methods for authorship attribution and verification. \colorbox{highlightorange}{\strut Text highlighted in orange} means that we obtain the log probabilities of the corresponding tokens, and \texttt{\{bracketed text\}} means variable interpolation. The top row defines common operations, while the bottom row defines the application of those operations to authorship verification and attribution. Random sampling (RS) and Zero-shot (ZS) are two different variants of our method for authorship verification.}
    \label{fig:method-overview}
\end{figure*}

\paragraph{Proposal.}
We propose a novel unsupervised approach to authorship attribution and verification that leverage the extensive Causal Language Modeling (CLM) pre-training, one-shot learning capabilities and scale of LLMs. For a given text $t$, we obtain a paraphrased version $N(t)$ in neutral style via LLM prompting. Next, we consider the task of predicting the styled text $t$ from its neutral version $N(t)$. LLMs can perform this task, especially with in-context examples. Given a single in-context example, the average of the LLM log-probabilities of the target text $t$ (or \emph{OSST score}) reflects how helpful the style of the in-context example was. Therefore, we use this metric to measure style similarity between the texts. The \emph{OSST score} can be used in both authorship attribution and verification. We provide a more in-depth explanation in \Cref{fig:method-overview} and \Cref{sec:method}.

Our results show OSST-based approaches consistently outperform comparably sized LLM prompting methods on authorship verification. With sufficiently large backbone models, they are also competitive with or surpass contrastive baselines across most of our experimental settings, and present good results in authorship attribution in non-English languages. We further observe a scaling behavior: performance increases with model size and, for the authorship verification, with an additional mechanism that increases test time compute.

\paragraph{Contributions.}
The main contributions of this work are the following.
\begin{enumerate}
    \item We proposed a novel unsupervised approach to authorship attribution and verification taking advantage of the extensive CLM pre-training, one-shot learning capabilities and scale of LLMs.
    \item We empirically validated its performance for authorship attribution and verification across several datasets and compared with contrastive and prompting-based baselines.
    \item We studied the performance of our methods across multiple base LLM sizes, and performed ablations on another verification-specific hyperparameter that scales test-time compute.
    \item We validated the multilingual performance of OSST-based methods in attribution.
\end{enumerate}

\section{Background}\label{sec:background}

\subsection{Authorship tasks}

In this work, we focus on two core authorship tasks: verification and attribution. Other authorship tasks, such as style change detection, can be reduced to them, and our method is not suited for authorship profiling (identifying features of the person who wrote a text) or clustering.

\paragraph{Authorship verification (AV).} Given a set of texts $T_a$ known to be written by the author $a$, the task is to make a binary decision on whether a target text $t$ was written by $a$.

\paragraph{Authorship attribution (AA).} Given a target text \(t\) and a set of candidate authors $A = \{a_1, a_2, \dots, a_n\}$, the task is to determine the most likely author of \(t\). To do this, a set of training texts $T = \{T_{a_1}, T_{a_2}, \dots, T_{a_n}\}$ is provided, where $T_{a_i}$ is a set of texts known to be written by the author $a_i$. In \emph{closed-set AA} the author of the text is always among the candidates. In \emph{open-set AA} this might not be the case, and methods must detect it.


\subsection{Available data}

Authorship datasets are foundational for advancing research in computational authorship analysis. Diverse corpora exist across multiple domains, including news articles~\cite{lewis2004RCV1new}, opinion pieces~\cite{stamatatosRobustnessAuthorshipAttribution2013}, product reviews~\cite{seroussiAuthorshipAttributionTopic2014}, blogs~\cite{schler2006Effectsage}, literary fiction~\cite{gerlach2020StandardizedProject}, and social media platforms such as Reddit~\cite{volskeTLDRMiningReddit2017} and Twitter~\cite{cheng2010Youare}. For evaluation, this work focuses on datasets that have been cleaned and standardized for AA and AV tasks.

A major driver of dataset quality and diversity in authorship analysis is the CLEF Initiative’s annual PAN competitions, which provide standardized benchmarks for digital text forensics. Early AA datasets (2011–2012) drew on corporate emails and fiction~\cite{argamon_2011_3713246,juola_2012_3713273}. The 2018 AA task introduced a cross-fandom FanFiction setting~\cite{kestemont_2018_3737849,kestemont2018overview}, and the 2019 edition extended this to an open-set scenario~\cite{kestemont_2019_3530313,kestemont2019overview}.

For AV, the 2020–2021 editions focused on FanFiction, with 2021 adding out-of-distribution authors and topics~\cite{bischoff2020ImportanceSuppressing,bevendorff2020PAN20Authorship}. In 2022, the task expanded to multiple discourse types (e.g., essays, emails, text messages, business memos)~\cite{stamatatos2022PAN22Authorship}.

We also consider style-change detection tasks (2022–2024), which require determining whether adjacent paragraphs share authorship. The 2022 dataset is based on StackExchange posts~\cite{zangerle2022PAN22Authorship}, while the 2023 and 2024 editions use topic-controlled Reddit data to reduce semantic cues and emphasize stylistic signals~\cite{zangerle2023PAN23MultiAuthor,zangerle2024PAN24MultiAuthor}.
\subsection{Content and style}

An open question in authorship is whether the style and the content of a text can be disentangled. Topic information is known to be predictive in some authorship datasets (see e.g.~\cite{sari-etal-2018-topic}). \citet{zangerleOverviewMultiauthorWriting2023} showed worse results in subtasks where topical correlation was accounted for.
Some work in the literature (e.g. ~\cite{wegmannSameAuthorJust2022,sawatpholTopicRegularizedAuthorshipRepresentation2022,patelStyleDistanceStrongerContentIndependent2025}) attempts to learn representations that are more independent of content. However, it is not clear that this can be done.
\citet{Mikros2007InvestigatingTI} studied several stylometric features that were topic-neutral in theory, and found many of them to be correlated with text topic. \citet{jafaritazehjaniStyleContentDistinction2020} found transferring stylistic features to a text while preserving content difficult, concluding content and style might be entangled.

\subsection{Authorship methods}\label{ssec:authorship-methods}

Traditional authorship analysis methods relied on handcrafted lexical (e.g., character and word n-grams, word frequency, word-length distribution), syntactic (e.g., POS distributions, syntactic n-grams), and semantic features~\cite{stamatatosAuthorshipVerificationReview2016,lagutinaSurveyStylometricText2019}. These features were typically used in supervised settings with classifiers such as SVMs~\cite{stamatatosAuthorshipVerificationReview2016}, or with early neural architectures including RNNs~\cite{bagnallAuthorIdentificationUsing2016} and CNNs~\cite{shrestha-etal-2017-convolutional}.

With the introduction of transformers~\cite{vaswani2017AttentionAll}, pre-trained encoder-only models such as BERT~\cite{devlin2019BERTPretraining} have been widely adopted for authorship-related tasks~\cite{fabien-etal-2020-bertaa,zangerleOverviewMultiauthorWriting2023}.

An alternative line of work fine-tunes an individual language model for each candidate author using a corpus of known texts, and attributes authorship by evaluating perplexity under these author-specific models~\cite{huangAttributingAuthorshipPerplexity2025,barlas2020CrossDomainAuthorship}. However, the need to fine-tune a language model for each author makes these methods expensive and not applicable in low-resource environments like authorship verification tasks.

There is a growing interest in learning author embeddings, mapping each text to a vector representation capturing stylistic properties. These methods still learn from supervised data, but rely on contrastive losses. They have the advantage of being more interpretable and simple to apply in all attribution tasks, including author profiling.
\citet{rivera-soto2021LearningUniversal} trained a convolutional model with contrastive learning.
\citet{wegmannSameAuthorJust2022} studied the use of the triplet contrastive loss with a hard negative from the same conversation to obtain embeddings that are more independent from content.
\citet{sawatpholTopicRegularizedAuthorshipRepresentation2022} used topic similarity (via a semantic similarity model) to modify the loss function and regularize training, reducing reliance on topical cues.
\citet{huertas-tatoUnderstandingWritingStyle2024} scaled training by efficiently implementing the use of large batches.
\citet{kimLeveragingMultilingualTraining2025} try to improve multilingual performance by using Language-Aware Batching (LAB), where examples in a batch are from the same language. They also used Probabilistic Content Masking (PCM) to randomly mask content words (words carrying meaning).
\citet{patelStyleDistanceStrongerContentIndependent2025} tried to reduce content-dependence by creating a synthetic dataset with LLMs. They create near-exact paraphrases with controlled style variations for 40 known style features.
\citet{agarwalCrossGenreAuthorshipAttribution2025} focused on cross-genre authorship attribution, trying to avoid topical cues. They implement hard negative mining in their contrastive pipeline, and curate a dataset to select topically dissimilar positive pairs.


A recent trend explores the use of large language models (LLMs) for human-like stylistic analysis based on prompts~\cite{hung2023WhoWrote,huang2024CanLarge}. They are promising unsupervised methods, but they require very large models in AA settings due to context limitations, and empirical evaluation remains scarce. Our results indicate that they yield limited performance in AV.

\subsection{Causal Language Modeling and in-context learning}

Causal Language Modeling (CLM) refers to predicting the next token based on prior tokens in a unidirectional, autoregressive manner, ie, modeling \(P(x_t \mid x_{<t})\)~\cite{bengio2003neuralprobabilistic}. It is the pre-training objective used by GPT-style decoder-only models~\cite{radford2019language}. A landmark discovery in large language models (LLMs) is their in-context learning ability~\cite{wei2022Emergentabilities}. GPT 3 (175B parameters) was shown to perform new tasks with just a few textual examples provided in the prompt, without any explicit gradient updates or fine-tuning, demonstrating competitive performance with fully fine-tuned systems~\cite{brown2020LanguageModels}.

\section{One-shot Style Transfer (OSST) LLM log-probabilities for authorship attribution and verification}\label{sec:method}

In this section we provide an in-depth explanation of the main components and intuitions behind our OSST-based methods for AA and AV. A high-level overview of all the components is given in \Cref{fig:method-overview}.

\subsection{General components}

\paragraph{Paraphrasing texts.}
Our methods begins by paraphrasing a given text in a neutral style. To do this, we prompt an instruction-tuned LLM using the template shown in \Cref{sec:neutralize-prompt}. More formally, given a text $t$, we consider the random variable $\mathrm{N}_{M,\theta}(t)$ of generating a neutral version with LLM $M$ and sampling parameters $\theta$. For the rest of the work, we omit the subscript and abuse the notation to signify a sample of the random variable.

The neutralization process merits discussion, since the existence of such an style is debatable and there is not a well defined process to obtain it. \emph{We note that we are not arguing for the existence of a neutral style or a neutralization process, only for the empirical usefulness of this line of prompting.} The intuition behind the design is to transform the text into a consistent and different anchor style while keeping as much semantic information as possible. This way, scores are comparable between different texts. We ablate other lines of prompting in \Cref{sec:ablation-paraphrasing-prompt}, comparing with simply asking the model to paraphrase or to rewrite in the style of Shakespeare. We find that the difference in performance between neutralizing and paraphrasing is not significant, while the Shakespeare style lags behind. It seems that the prompt choice is not irrelevant, but other sensitive choices may also be acceptable.

\paragraph{One-shot style transfer LLM log-probabilities.}
Our method relies on the one-shot capabilities of LLMs to transfer the style of one text to another. In particular, for a pair of texts $t_1$ and $t_2$, consider the following input:

\begin{center}
\begin{tcolorbox}[width=0.7\linewidth, boxrule=0.5pt, fontupper=\small]
Neutral text: \textcolor{orange}{\{$\mathrm{N}(t_1)$\}}\\
Styled text: \textcolor{orange}{\{$t_1$\}}\\[0.5em]

Neutral text: \textcolor{orange}{\{$\mathrm{N}(t_2)$\}}\\
Styled text: \textcolor{orange}{\{$t_2$\}}
\end{tcolorbox}
\end{center}

We frame the problem as a style transfer task from the neutral version to the styled version of the second text, based on the style of the one-shot example. In this way, the LLM can predict semantic information from the neutral version while obtaining stylistic information from the one-shot example.
To measure the degree of transferability from the style of the first text to the second text, we retrieve the log-probabilities from a given \emph{foundational LLM}:
\begin{equation}
\mathrm{OSST}(t_2|t_1) = \log P \left[ t_2 \mid \mathrm{N}(t_2);\  t_1, \mathrm{N}(t_1)\right]
.\end{equation}
In other words, the model is measuring if the second text could come from transferring the style of the first text to the neutral version of the second text.

\subsection{Variability and normalization of the OSST score}\label{ssec:osst-var}

We observe that the OSST score defined above may have a varying distribution depending on the target text $t_2$ (its length, topic, genre...). For example, initial tokens are harder to predict because there is limited preceding context within the styled target text (only the neutral version and one-shot example are available). Thus, shorter target texts will naturally have a higher OSST score. Therefore, we want to \emph{normalize} these scores if we are to compare scores obtained for different target texts.


\subsection{Authorship attribution}

\paragraph{Closed-set AA.} Given a target text \(t_{target}\), a set of candidate authors $\mathcal{A} = \{a_1, a_2, \dots, a_n\}$, and a set of training texts $\mathcal T = \{T_{a_1}, T_{a_2}, \dots, T_{a_n}\}$, our method assigns a score to each candidate author as
\begin{equation}
\mathrm{score}(a) = \mathop{\mathrm{mean}}_{t \in T_{a}} 
    \ \mathrm{OSST}(t_{target}|t)
,\end{equation}
and the prediction is the candidate with the highest score. Since we only compare scores for the given target text, we do not need any form of normalization in this setting as per \Cref{ssec:osst-var}.

\paragraph{Open-set AA.} The open-set attribution setting includes the possibility that the target text is not written by any of the candidate authors. Unlike in the closed-set setting, we must select a decision boundary such that \emph{if the highest score of the candidates is below this threshold, then we predict that no candidate is the author}. This boundary must be selected globally for all inputs, and this requires comparing OSST scores for different target texts. Therefore, we want to apply some form of normalization as per \Cref{ssec:osst-var}.

The OSST score is computed for the same target text across several candidate texts, providing a simple way to \emph{standardize the scores} without additional computations:
\begin{equation}
\begin{cases}
\mu(t_{target}) = \mathop{\mathrm{mean}}\limits_{a\in \mathcal{A}, t \in T_{a}} 
    \ \mathrm{OSST}(t_{target}|t), \\
\sigma(t_{target}) = \mathop{\mathrm{std}}\limits_{a\in \mathcal{A}, t \in T_{a}} 
    \ \mathrm{OSST}(t_{target}|t). \\
\end{cases}
\end{equation}
This alleviates the natural discrepancies between target texts and makes the scores more comparable.

\subsection{Authorship verification}

We consider the task deciding whether two texts $t_1$ and $t_2$ were authored by the same person. Since the task is symmetric across texts, we can consider the score
\begin{multline}
    \mathrm{score}(t_1,t_2) =\\
    \frac{1}{2} \left( \mathrm{OSST}(t_1|t_2) + \mathrm{OSST}(t_2|t_1)\right)
.\end{multline}
AV presents the same difficulty as open-set AA. We have to define a global decision boundary such that a pair is predicted to be by the same author if their OSST score exceeds this threshold. \emph{This requires some form of normalization} as per \Cref{ssec:osst-var}. But unlike open-set AA, we only have one OSST score for each target text. Thus, we have to produce other OSST (or OSST-like) scores for the same target text to normalize. We present two options to solve this problem.

\begin{description}
\item[Zero-shot baseline (nOSST-ZS).] We \textit{normalize} the \(\mathrm{OSST}\) score by comparing it with the log-probabilities of the zero-shot problem, without any in-context examples. This is shown graphically in \Cref{fig:method-overview}, and mathematically as follows:
\begin{multline}
\mathrm{nOSST}_{zs}(t_2|t_1) = \mathrm{OSST}(t_2|t_1) \\
- \log P \left[ t_2 \mid \mathrm{N}(t_2)\right]
\end{multline}
 
This normalization technique has some problems. \Cref{fig:noost-positions} shows the median $\mathrm{nOSST}_{zs}$ score (without averaging across tokens of the target text) by token position in the PAN23 style change subsets. As we can see, the first few tokens have a much higher median value, and it progressively stabilizes. This still presents a challenge in comparing texts of different lengths, because shorter texts will have a higher overall score.

\begin{figure}[!ht]
    \centering
    \includegraphics[width=\linewidth]{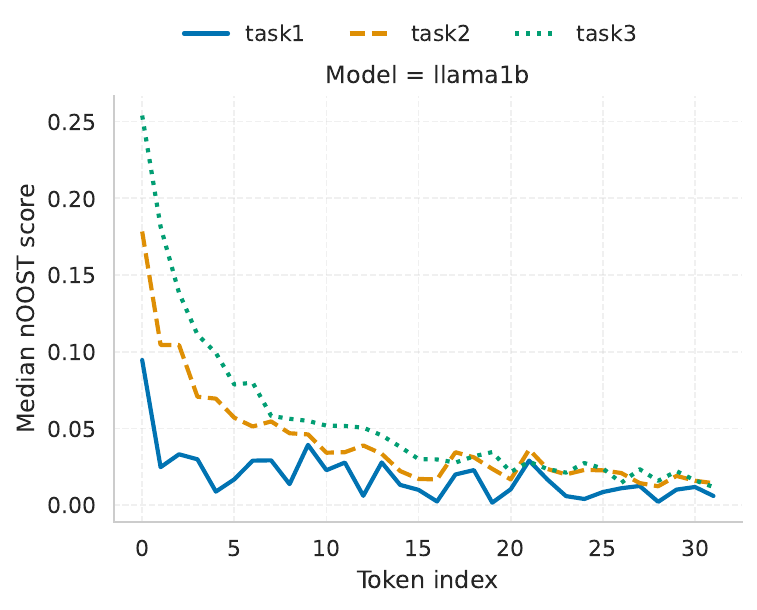}
    \caption{Median $\mathrm{nOSST}_{zs}$ score (without averaging across tokens of the target text) by token position in the PAN23 test sets.}
    \label{fig:noost-positions}
\end{figure}

\item[Random sampling baseline (nOSST-RS).] To mitigate the previous problems, we can select a fixed set of \emph{anchor texts} \(\mathcal{T}_{anchor}\), and compute the normalized score as
\begin{multline}
\mathrm{nOSST}_{rs}(t_2|t_1) = \mathrm{OSST}(t_2|t_1) - \\
\mathop{\mathrm{mean}}_{t \in T_{anchor}}  \mathrm{OSST}(t_2|t)
.\end{multline}
In this case, the difficulty lies in sampling appropriate anchor texts. In this work, we sample in-distribution from the corresponding training set. This is a limitation, as it is not feasible in real-world applications. We may be able to use contrastive models to sample in-distribution anchors, but this is not studied here.
\end{description}

\section{Experimental results}\label{sec:results}

We conducted several experiments with the purpose of evaluating and studying OSST-based methods in attribution and verification. Since OSST-based methods are unsupervised, we focus on baselines that can be applied without supervised training, such as contrastive and prompt-based methods.

\subsection{Authorship attribution}\label{ssec:results-attribution}


We evaluated our method in the fiction domain with the PAN12~\cite{juola_2012_3713273}, PAN18~\cite{kestemont_2018_3737849}, and PAN19~\cite{kestemont_2019_3530313} attribution datasets, and in the corporate email domain with PAN11~\cite{argamon_2011_3713246}. As contrastive learning baselines we use STAR~\cite{huertas-tatoUnderstandingWritingStyle2024}, LUAR~\cite{rivera-soto2021LearningUniversal}, StyleEmbedding~\cite{wegmannSameAuthorJust2022}, StyleDistance~\cite{patelStyleDistanceStrongerContentIndependent2025} and the multilingual models from~\cite{kimLeveragingMultilingualTraining2025} (with XLM-RoBERTa and Llama3.2-1B backbones). We do not compare against LLM prompting approaches such as LIP~\cite{huang2024CanLarge}, as these methods require large context windows in AA, resulting in markedly lower performance at moderate model sizes. Finally, OSST experiments are run three times.

\Cref{tab:results-attribution-closed} presents the results in closed-set attribution. In the fiction domain, our method with 4B and 8B parameters outperforms the baselines across all datasets, with the smaller Llama1B backbone performing competitively with all baselines except STAR. There is an increase in performance with larger model sizes, suggesting that further gains may be achievable with continued scaling. In the PAN11 dataset, all OSST models achieve higher accuracy but lower rank than some of the baselines. Interestingly, model performance does not improve with model size in PAN11. A likely explanation is its noisier nature compared to other datasets, as it contains erroneous messages, irregular candidate set sizes, and other inconsistencies.

\begin{table*}[!ht]
    \centering
    \caption{Closed-set authorship attribution results. Accuracy and average rank of correct candidate are reported.}
    \label{tab:results-attribution-closed}
    \centering
    \begin{adjustbox}{width=\textwidth}
    \begin{tabular}{l|cc|cc|cc|cc}
\toprule
 & \multicolumn{2}{|c}{PAN11} & \multicolumn{2}{|c}{PAN12} & \multicolumn{2}{|c}{PAN18-en} & \multicolumn{2}{|c}{PAN19-en} \\
 & Accuracy & Mean rank & Accuracy & Mean rank & Accuracy & Mean rank & Accuracy & Mean rank \\
\midrule
OSST-Gemma1B & \databar{0.23714020427112348}{0.8900203665987779}{2}{pastel2}{\textbf{23.71\%}\textsubscript{±0.23\%}} & \databar{0.8194827353601726}{0.97333484197028}{2}{pastel3}{0.82\textsubscript{±0.00}} & \databar{0.3333333333333333}{0.8900203665987779}{2}{pastel2}{33.33\%\textsubscript{±1.03\%}} & \databar{0.7033375850340136}{0.97333484197028}{2}{pastel3}{0.70\textsubscript{±0.03}} & \databar{0.4641148325358852}{0.8900203665987779}{2}{pastel2}{46.41\%\textsubscript{±1.66\%}} & \databar{0.8403774587985114}{0.97333484197028}{2}{pastel3}{0.84\textsubscript{±0.01}} & \databar{0.6995926680244399}{0.8900203665987779}{2}{pastel2}{69.96\%\textsubscript{±1.24\%}} & \databar{0.9032586558044806}{0.97333484197028}{2}{pastel3}{0.90\textsubscript{±0.00}} \\
OSST-Llama1B & \databar{0.2263695450324977}{0.8900203665987779}{2}{pastel2}{22.64\%\textsubscript{±0.20\%}} & \databar{0.8278478124578402}{0.97333484197028}{2}{pastel3}{0.83\textsubscript{±0.00}} & \databar{0.44047619047619047}{0.8900203665987779}{2}{pastel2}{44.05\%\textsubscript{±8.25\%}} & \databar{0.7643140589569161}{0.97333484197028}{2}{pastel3}{0.76\textsubscript{±0.04}} & \databar{0.521531100478469}{0.8900203665987779}{2}{pastel2}{52.15\%\textsubscript{±2.09\%}} & \databar{0.8576289207868156}{0.97333484197028}{2}{pastel3}{0.86\textsubscript{±0.02}} & \databar{0.7240325865580447}{0.8900203665987779}{2}{pastel2}{72.40\%\textsubscript{±0.88\%}} & \databar{0.9119710341706267}{0.97333484197028}{2}{pastel3}{0.91\textsubscript{±0.01}} \\
OSST-Gemma4B & \databar{0.23714020427112348}{0.8900203665987779}{2}{pastel2}{\textbf{23.71\%}\textsubscript{±0.49\%}} & \databar{0.8267935226848876}{0.97333484197028}{2}{pastel3}{0.83\textsubscript{±0.00}} & \databar{0.6071428571428571}{0.8900203665987779}{2}{pastel2}{60.71\%\textsubscript{±3.09\%}} & \databar{0.866000566893424}{0.97333484197028}{2}{pastel3}{0.87\textsubscript{±0.02}} & \databar{0.6331738437001595}{0.8900203665987779}{2}{pastel2}{63.32\%\textsubscript{±1.54\%}} & \databar{0.932695374800638}{0.97333484197028}{2}{pastel3}{0.93\textsubscript{±0.01}} & \databar{0.8448744059742023}{0.8900203665987779}{2}{pastel2}{84.49\%\textsubscript{±0.24\%}} & \databar{0.960322848306555}{0.97333484197028}{2}{pastel3}{0.96\textsubscript{±0.00}} \\
OSST-Llama8B & \databar{0.21448467966573817}{0.8900203665987779}{2}{pastel2}{21.45\%\textsubscript{±0.29\%}} & \databar{0.8271964700934061}{0.97333484197028}{2}{pastel3}{0.83\textsubscript{±0.00}} & \databar{0.7142857142857143}{0.8900203665987779}{2}{pastel2}{\textbf{71.43\%}\textsubscript{±3.09\%}} & \databar{0.9400510204081632}{0.97333484197028}{2}{pastel3}{\textbf{0.94}\textsubscript{±0.01}} & \databar{0.7049441786283891}{0.8900203665987779}{2}{pastel2}{\textbf{70.49\%}\textsubscript{±1.10\%}} & \databar{0.9482987772461456}{0.97333484197028}{2}{pastel3}{\textbf{0.95}\textsubscript{±0.00}} & \databar{0.8900203665987779}{0.8900203665987779}{2}{pastel2}{\textbf{89.00\%}\textsubscript{±0.71\%}} & \databar{0.97333484197028}{0.97333484197028}{2}{pastel3}{\textbf{0.97}\textsubscript{±0.00}} \\
\midrule
STAR & \databar{0.17827298050139276}{0.8879837067209776}{2}{pastel2}{17.83\%} & \databar{0.8536140275694593}{0.9723919438787056}{2}{pastel3}{\textbf{0.85}} & \databar{0.5714285714285714}{0.8879837067209776}{2}{pastel2}{57.14\%} & \databar{0.8687287414965985}{0.9723919438787056}{2}{pastel3}{0.87} & \databar{0.6267942583732058}{0.8879837067209776}{2}{pastel2}{62.68\%} & \databar{0.9255980861244019}{0.9723919438787056}{2}{pastel3}{0.93} & \databar{0.7606924643584522}{0.8879837067209776}{2}{pastel2}{76.07\%} & \databar{0.9353926227653316}{0.9723919438787056}{2}{pastel3}{0.94} \\
LUAR$_{CRUD}$ & \databar{0.16211699164345403}{0.8879837067209776}{2}{pastel2}{16.21\%} & \databar{0.8368687950860653}{0.9723919438787056}{2}{pastel3}{0.84} & \databar{0.39285714285714285}{0.8879837067209776}{2}{pastel2}{39.29\%} & \databar{0.7631802721088435}{0.9723919438787056}{2}{pastel3}{0.76} & \databar{0.4784688995215311}{0.8879837067209776}{2}{pastel2}{47.85\%} & \databar{0.8768740031897927}{0.9723919438787056}{2}{pastel3}{0.88} & \databar{0.6955193482688391}{0.8879837067209776}{2}{pastel2}{69.55\%} & \databar{0.9143471373613941}{0.9723919438787056}{2}{pastel3}{0.91} \\
LUAR$_{MUD}$ & \databar{0.1799442896935933}{0.8879837067209776}{2}{pastel2}{17.99\%} & \databar{0.8512469347427565}{0.9723919438787056}{2}{pastel3}{0.85} & \databar{0.42857142857142855}{0.8879837067209776}{2}{pastel2}{42.86\%} & \databar{0.7816751700680271}{0.9723919438787056}{2}{pastel3}{0.78} & \databar{0.5454545454545454}{0.8879837067209776}{2}{pastel2}{54.55\%} & \databar{0.8811004784688995}{0.9723919438787056}{2}{pastel3}{0.88} & \databar{0.620162932790224}{0.8879837067209776}{2}{pastel2}{62.02\%} & \databar{0.9062004978501924}{0.9723919438787056}{2}{pastel3}{0.91} \\
StyEmb & \databar{0.06685236768802229}{0.8879837067209776}{2}{pastel2}{6.69\%} & \databar{0.7441075637454466}{0.9723919438787056}{2}{pastel3}{0.74} & \databar{0.26785714285714285}{0.8879837067209776}{2}{pastel2}{26.79\%} & \databar{0.7224702380952381}{0.9723919438787056}{2}{pastel3}{0.72} & \databar{0.5454545454545454}{0.8879837067209776}{2}{pastel2}{54.55\%} & \databar{0.8760765550239235}{0.9723919438787056}{2}{pastel3}{0.88} & \databar{0.5804480651731161}{0.8879837067209776}{2}{pastel2}{58.04\%} & \databar{0.8622991627064948}{0.9723919438787056}{2}{pastel3}{0.86} \\
StyDist & \databar{0.10473537604456824}{0.8879837067209776}{2}{pastel2}{10.47\%} & \databar{0.7351171344903935}{0.9723919438787056}{2}{pastel3}{0.74} & \databar{0.375}{0.8879837067209776}{2}{pastel2}{37.50\%} & \databar{0.7443664965986395}{0.9723919438787056}{2}{pastel3}{0.74} & \databar{0.48325358851674644}{0.8879837067209776}{2}{pastel2}{48.33\%} & \databar{0.8668261562998406}{0.9723919438787056}{2}{pastel3}{0.87} & \databar{0.6008146639511202}{0.8879837067209776}{2}{pastel2}{60.08\%} & \databar{0.8668250735460511}{0.9723919438787056}{2}{pastel3}{0.87} \\
MLangRob & \databar{0.14317548746518105}{0.8879837067209776}{2}{pastel2}{14.32\%} & \databar{0.8182195319381949}{0.9723919438787056}{2}{pastel3}{0.82} & \databar{0.4642857142857143}{0.8879837067209776}{2}{pastel2}{46.43\%} & \databar{0.7895408163265306}{0.9723919438787056}{2}{pastel3}{0.79} & \databar{0.49760765550239233}{0.8879837067209776}{2}{pastel2}{49.76\%} & \databar{0.880781499202552}{0.9723919438787056}{2}{pastel3}{0.88} & \databar{0.45010183299389}{0.8879837067209776}{2}{pastel2}{45.01\%} & \databar{0.8086671192577506}{0.9723919438787056}{2}{pastel3}{0.81} \\
MLangLlama & \databar{0.19832869080779944}{0.8879837067209776}{2}{pastel2}{19.83\%} & \databar{0.8075548175130348}{0.9723919438787056}{2}{pastel3}{0.81} & \databar{0.39285714285714285}{0.8879837067209776}{2}{pastel2}{39.29\%} & \databar{0.7899659863945577}{0.9723919438787056}{2}{pastel3}{0.79} & \databar{0.47368421052631576}{0.8879837067209776}{2}{pastel2}{47.37\%} & \databar{0.8384370015948964}{0.9723919438787056}{2}{pastel3}{0.84} & \databar{0.4164969450101833}{0.8879837067209776}{2}{pastel2}{41.65\%} & \databar{0.8168137587689522}{0.9723919438787056}{2}{pastel3}{0.82} \\
\bottomrule
\end{tabular}
    \end{adjustbox}
\end{table*}


\Cref{tab:results-attribution-open} summarizes results in the open-set setting on the PAN19 dataset. A small held-out subset is used to select the decision threshold for whether the true author is among the candidates. Scores are standardized across candidate texts (for all methods) to improve comparability of scores for different target texts. Our method exhibits better performance in the open-set scenario than contrastive baselines with all parameter counts.

\begin{table}[!ht]
\caption{Open-set authorship attribution evaluation in PAN19. BinF1 refers to the binary classification of true author presence among candidates, while the rank of the correct candidate is calculated only for instances where the true author is included. All similarity scores are standardized across candidate texts.}
\label{tab:results-attribution-open}
\centering
\begin{adjustbox}{width=\linewidth}

\begin{tabular}{lccc}
\toprule
 & Accuracy & BinF1Macro & Mean rank \\
\midrule
OSST-Gemma1B & \databar{0.631004366812227}{0.7612809315866085}{2}{pastel2}{63.10\%\textsubscript{±0.58\%}} & \databar{0.6458979841197056}{0.7546120449124407}{2}{pastel1}{64.59\%\textsubscript{±1.60\%}} & \databar{0.9031155619138452}{0.9718049594659037}{2}{pastel3}{0.90\textsubscript{±0.00}} \\
OSST-Llama1B & \databar{0.6482290150412421}{0.7612809315866085}{2}{pastel2}{64.82\%\textsubscript{±0.08\%}} & \databar{0.6704948387020554}{0.7546120449124407}{2}{pastel1}{67.05\%\textsubscript{±0.14\%}} & \databar{0.9139246542680018}{0.9718049594659037}{2}{pastel3}{0.91\textsubscript{±0.01}} \\
OSST-Gemma4B & \databar{0.7428432799611838}{0.7612809315866085}{2}{pastel2}{74.28\%\textsubscript{±0.67\%}} & \databar{0.7490986364718281}{0.7546120449124407}{2}{pastel1}{74.91\%\textsubscript{±0.72\%}} & \databar{0.9604991257351774}{0.9718049594659037}{2}{pastel3}{0.96\textsubscript{±0.00}} \\
OSST-Llama8B & \databar{0.7612809315866085}{0.7612809315866085}{2}{pastel2}{\textbf{76.13\%}\textsubscript{±1.96\%}} & \databar{0.7546120449124407}{0.7546120449124407}{2}{pastel1}{\textbf{75.46\%}\textsubscript{±2.40\%}} & \databar{0.9718049594659037}{0.9718049594659037}{2}{pastel3}{\textbf{0.97}\textsubscript{±0.00}} \\
\midrule
STAR & \databar{0.5938864628820961}{0.7612809315866085}{2}{pastel2}{59.39\%} & \databar{0.6136477477641573}{0.7546120449124407}{2}{pastel1}{61.36\%} & \databar{0.9347877920839294}{0.9718049594659037}{2}{pastel3}{0.93} \\
LUAR$_{CRUD}$ & \databar{0.606259097525473}{0.7612809315866085}{2}{pastel2}{60.63\%} & \databar{0.6551994364708991}{0.7546120449124407}{2}{pastel1}{65.52\%} & \databar{0.9160705770147831}{0.9718049594659037}{2}{pastel3}{0.92} \\
LUAR$_{MUD}$ & \databar{0.5312954876273653}{0.7612809315866085}{2}{pastel2}{53.13\%} & \databar{0.6206541178574623}{0.7546120449124407}{2}{pastel1}{62.07\%} & \databar{0.9067715784453981}{0.9718049594659037}{2}{pastel3}{0.91} \\
StyEmb& \databar{0.5305676855895196}{0.7612809315866085}{2}{pastel2}{53.06\%} & \databar{0.5838959916049562}{0.7546120449124407}{2}{pastel1}{58.39\%} & \databar{0.8619456366237482}{0.9718049594659037}{2}{pastel3}{0.86} \\
StyDist& \databar{0.517467248908297}{0.7612809315866085}{2}{pastel2}{51.75\%} & \databar{0.589845011055874}{0.7546120449124407}{2}{pastel1}{58.98\%} & \databar{0.86552217453505}{0.9718049594659037}{2}{pastel3}{0.87} \\
MLangRob & \databar{0.3195050946142649}{0.7612809315866085}{2}{pastel2}{31.95\%} & \databar{0.45028432873606344}{0.7546120449124407}{2}{pastel1}{45.03\%} & \databar{0.8096089651883643}{0.9718049594659037}{2}{pastel3}{0.81} \\
MLangLlama & \databar{0.42212518195050946}{0.7612809315866085}{2}{pastel2}{42.21\%} & \databar{0.5676180693129845}{0.7546120449124407}{2}{pastel1}{56.76\%} & \databar{0.8192656175488794}{0.9718049594659037}{2}{pastel3}{0.82} \\
\bottomrule
\end{tabular}
\end{adjustbox}
\end{table}
\subsection{Authorship verification}\label{ssec:results-verification}

For authorship verification (AV), we evaluate on the 2020 and 2021 editions of the PAN AV task~\cite{bischoff2020ImportanceSuppressing, bevendorff2020PAN20Authorship}, which focus on FanFiction, and in the PAN style-change detection tasks of 2022–2024~\cite{zangerle2022PAN22Authorship, zangerle2023PAN23MultiAuthor, zangerle2024PAN24MultiAuthor}. The 2022 edition contains StackExchange posts, while the 2023 and 2024 editions comprise Reddit posts and explicitly control for topic in the second and third tasks, thereby preventing detectors from exploiting semantic information.

We use the same constrastive baselines as in AA. We include LIP~\cite{huang2024CanLarge} and PromptAV~\cite{hung2023WhoWrote} with a 24B Mistral LLM as unsupervised baselines. For contrastive and OSST-based methods, we select a decision boundary with 2000 examples from the training set. For nOSST methods, we work with the LLaMA base models and a 24B Mistral model to test larger-capacity base models. \emph{In the nOSST-RS setting, we use 10 anchors sampled in-distribution from the training set}.

The results on authorship verification datasets are summarized in \Cref{tab:results-verification}. We make several observations:
\begin{enumerate*}[label=(\roman*)]
\item In nOSST methods, larger backbone models tend to achieve higher accuracy.
\item nOSST-ZS with the larger backbone models performs competitively with contrastive baselines on average and in the harder PAN23SC-T3 and PAN24SC-T3 tasks where there is less topical correlation.
\item nOSST-RS generally improves performance with respect to nOSST-ZS across multiple datasets, though a few exceptions remain.
\item nOSST-RS with a Mistral24B backbone is the best overall, and it pushes performance on the more difficult sets.
\item At comparable model scales (up to 24B parameters), nOSST methods outperform LLM prompting approaches.
\end{enumerate*}



\begin{table*}[!ht]
\caption{Results on authorship verification tasks. F1 score is reported. Best result per dataset is marked in bold.}
\label{tab:results-verification}
\centering
\begin{adjustbox}{width=\textwidth}

\begin{tabular}{l|ccc|ccc|ccc|cc|c}
\toprule
 & \multicolumn{3}{c|}{22ST} & \multicolumn{3}{c|}{23ST} & \multicolumn{3}{c|}{24ST} & 20AV & 21AV & Avg\\
 & T1 & T2 & T3 & T1 & T2 & T3 & T1 & T2 & T3 &  &  & \\
\midrule
nOSST$_{ZS}^{1B}$ & \databar{0.6226331730419634}{0.8455677672863762}{1.1}{pastel1}{62.26\%} & \databar{0.6257111777048504}{0.8455677672863762}{1.1}{pastel1}{62.57\%} & \databar{0.5830122673300138}{0.8455677672863762}{1.1}{pastel1}{58.30\%} & \databar{0.6900613770375695}{0.8455677672863762}{1.1}{pastel1}{69.01\%} & \databar{0.5582210180460465}{0.8455677672863762}{1.1}{pastel1}{55.82\%} & \databar{0.5896835207515678}{0.8455677672863762}{1.1}{pastel1}{58.97\%} & \databar{0.7320525793725154}{0.8455677672863762}{1.1}{pastel1}{73.21\%} & \databar{0.6712742312494779}{0.8455677672863762}{1.1}{pastel1}{67.13\%} & \databar{0.6126927172805838}{0.8455677672863762}{1.1}{pastel1}{61.27\%} & \databar{0.6764963913709701}{0.8455677672863762}{1.1}{pastel1}{67.65\%} & \databar{0.6733727946355741}{0.8455677672863762}{1.1}{pastel1}{67.34\%} & \databar{0.6395646588928302}{0.8455677672863762}{1.1}{pastel1}{63.96\%} \\
nOSST$_{ZS}^{8B}$ & \databar{0.6457309693608932}{0.8455677672863762}{1.1}{pastel1}{64.57\%} & \databar{0.6687141004777233}{0.8455677672863762}{1.1}{pastel1}{66.87\%} & \databar{0.5998814102036238}{0.8455677672863762}{1.1}{pastel1}{59.99\%} & \databar{0.6863386423808745}{0.8455677672863762}{1.1}{pastel1}{68.63\%} & \databar{0.6153526176950219}{0.8455677672863762}{1.1}{pastel1}{61.54\%} & \databar{0.6292736136588788}{0.8455677672863762}{1.1}{pastel1}{62.93\%} & \databar{0.7541517665290601}{0.8455677672863762}{1.1}{pastel1}{75.42\%} & \databar{0.6927772177962457}{0.8455677672863762}{1.1}{pastel1}{69.28\%} & \databar{0.6335776743502647}{0.8455677672863762}{1.1}{pastel1}{63.36\%} & \databar{0.7621592388972664}{0.8455677672863762}{1.1}{pastel1}{76.22\%} & \databar{0.7650142117987493}{0.8455677672863762}{1.1}{pastel1}{76.50\%} & \databar{0.6775428602862366}{0.8455677672863762}{1.1}{pastel1}{67.75\%} \\
nOSST$_{ZS}^{24B}$ & \databar{0.6638472693139031}{0.8455677672863762}{1.1}{pastel1}{66.38\%} & \databar{0.6954767527602513}{0.8455677672863762}{1.1}{pastel1}{69.55\%} & \databar{0.6201814690065591}{0.8455677672863762}{1.1}{pastel1}{62.02\%} & \databar{0.714888092568158}{0.8455677672863762}{1.1}{pastel1}{71.49\%} & \databar{0.6305530076182964}{0.8455677672863762}{1.1}{pastel1}{63.06\%} & \databar{0.6454120297996778}{0.8455677672863762}{1.1}{pastel1}{64.54\%} & \databar{0.7655641052495394}{0.8455677672863762}{1.1}{pastel1}{76.56\%} & \databar{0.7583936699024336}{0.8455677672863762}{1.1}{pastel1}{75.84\%} & \databar{0.6572578247504737}{0.8455677672863762}{1.1}{pastel1}{65.73\%} & \databar{0.7918085919650659}{0.8455677672863762}{1.1}{pastel1}{79.18\%} & \databar{0.7929804936467937}{0.8455677672863762}{1.1}{pastel1}{79.30\%} & \databar{0.7033057551437412}{0.8455677672863762}{1.1}{pastel1}{70.33\%} \\
nOSST$_{RS}^{1B}$ & \databar{0.6354826084937679}{0.8455677672863762}{1.1}{pastel1}{63.55\%} & \databar{0.6807597703635463}{0.8455677672863762}{1.1}{pastel1}{68.08\%} & \databar{0.6131907840130573}{0.8455677672863762}{1.1}{pastel1}{61.32\%} & \databar{0.7378820367933298}{0.8455677672863762}{1.1}{pastel1}{73.79\%} & \databar{0.6138254080215129}{0.8455677672863762}{1.1}{pastel1}{61.38\%} & \databar{0.6193280484109156}{0.8455677672863762}{1.1}{pastel1}{61.93\%} & \databar{0.8103209303130163}{0.8455677672863762}{1.1}{pastel1}{81.03\%} & \databar{0.6198906691489285}{0.8455677672863762}{1.1}{pastel1}{61.99\%} & \databar{0.6301861560154018}{0.8455677672863762}{1.1}{pastel1}{63.02\%} & \databar{0.7322234424968904}{0.8455677672863762}{1.1}{pastel1}{73.22\%} & \databar{0.7237211142941686}{0.8455677672863762}{1.1}{pastel1}{72.37\%} & \databar{0.6742555425785941}{0.8455677672863762}{1.1}{pastel1}{67.43\%} \\
nOSST$_{RS}^{8B}$ & \databar{0.6516351644222576}{0.8455677672863762}{1.1}{pastel1}{65.16\%} & \databar{0.7145803069496923}{0.8455677672863762}{1.1}{pastel1}{71.46\%} & \databar{0.6223428213440925}{0.8455677672863762}{1.1}{pastel1}{62.23\%} & \databar{0.7622530474989492}{0.8455677672863762}{1.1}{pastel1}{76.23\%} & \databar{0.7025376096768032}{0.8455677672863762}{1.1}{pastel1}{70.25\%} & \databar{0.6667347256004132}{0.8455677672863762}{1.1}{pastel1}{66.67\%} & \databar{0.8306789136772068}{0.8455677672863762}{1.1}{pastel1}{83.07\%} & \databar{0.685840395480226}{0.8455677672863762}{1.1}{pastel1}{68.58\%} & \databar{0.6616857763245141}{0.8455677672863762}{1.1}{pastel1}{66.17\%} & \databar{0.8083558789179619}{0.8455677672863762}{1.1}{pastel1}{80.84\%} & \databar{0.8001495521258538}{0.8455677672863762}{1.1}{pastel1}{80.01\%} & \databar{0.7187994720016337}{0.8455677672863762}{1.1}{pastel1}{71.88\%} \\
nOSST$_{RS}^{24B}$ & \databar{0.6772541325523725}{0.8455677672863762}{1.1}{pastel1}{\textbf{67.73\%}} & \databar{0.7498559738139978}{0.8455677672863762}{1.1}{pastel1}{\textbf{74.99\%}} & \databar{0.6410476563925317}{0.8455677672863762}{1.1}{pastel1}{\textbf{64.10\%}} & \databar{0.7775616460788248}{0.8455677672863762}{1.1}{pastel1}{77.76\%} & \databar{0.6982991797151632}{0.8455677672863762}{1.1}{pastel1}{69.83\%} & \databar{0.6830151598326915}{0.8455677672863762}{1.1}{pastel1}{\textbf{68.30\%}} & \databar{0.8441531820959063}{0.8455677672863762}{1.1}{pastel1}{\textbf{84.42\%}} & \databar{0.8000161511749979}{0.8455677672863762}{1.1}{pastel1}{\textbf{80.00\%}} & \databar{0.6866213351137974}{0.8455677672863762}{1.1}{pastel1}{\textbf{68.66\%}} & \databar{0.8455677672863762}{0.8455677672863762}{1.1}{pastel1}{\textbf{84.56\%}} & \databar{0.8349770111482103}{0.8455677672863762}{1.1}{pastel1}{\textbf{83.50\%}} & \databar{0.7489426541095336}{0.8455677672863762}{1.1}{pastel1}{\textbf{74.89\%}} \\
\midrule
STAR & \databar{0.6461611913987183}{0.8455677672863762}{1.1}{pastel1}{64.62\%} & \databar{0.6685438056009798}{0.8455677672863762}{1.1}{pastel1}{66.85\%} & \databar{0.5805938788186862}{0.8455677672863762}{1.1}{pastel1}{58.06\%} & \databar{0.794583593217691}{0.8455677672863762}{1.1}{pastel1}{79.46\%} & \databar{0.728555929917769}{0.8455677672863762}{1.1}{pastel1}{72.86\%} & \databar{0.5977386634615014}{0.8455677672863762}{1.1}{pastel1}{59.77\%} & \databar{0.7979216568316376}{0.8455677672863762}{1.1}{pastel1}{79.79\%} & \databar{0.768682556953628}{0.8455677672863762}{1.1}{pastel1}{76.87\%} & \databar{0.5796519547219676}{0.8455677672863762}{1.1}{pastel1}{57.97\%} & \databar{0.7656875268868768}{0.8455677672863762}{1.1}{pastel1}{76.57\%} & \databar{0.7922361191204842}{0.8455677672863762}{1.1}{pastel1}{79.22\%} & \databar{0.701850625175449}{0.8455677672863762}{1.1}{pastel1}{70.19\%} \\
LUAR$_{CRUD}$ & \databar{0.667809640992306}{0.8455677672863762}{1.1}{pastel1}{66.78\%} & \databar{0.7066983837123851}{0.8455677672863762}{1.1}{pastel1}{70.67\%} & \databar{0.6240734628526202}{0.8455677672863762}{1.1}{pastel1}{62.41\%} & \databar{0.8075689353626316}{0.8455677672863762}{1.1}{pastel1}{80.76\%} & \databar{0.7365900886915577}{0.8455677672863762}{1.1}{pastel1}{\textbf{73.66\%}} & \databar{0.622454814856894}{0.8455677672863762}{1.1}{pastel1}{62.25\%} & \databar{0.7831657277242493}{0.8455677672863762}{1.1}{pastel1}{78.32\%} & \databar{0.7636024065780024}{0.8455677672863762}{1.1}{pastel1}{76.36\%} & \databar{0.6010284906551197}{0.8455677672863762}{1.1}{pastel1}{60.10\%} & \databar{0.6874259234296275}{0.8455677672863762}{1.1}{pastel1}{68.74\%} & \databar{0.699752161940736}{0.8455677672863762}{1.1}{pastel1}{69.98\%} & \databar{0.7000154578905572}{0.8455677672863762}{1.1}{pastel1}{70.00\%} \\
LUAR$_{MUD}$ & \databar{0.6666302488911218}{0.8455677672863762}{1.1}{pastel1}{66.66\%} & \databar{0.6985408218665405}{0.8455677672863762}{1.1}{pastel1}{69.85\%} & \databar{0.6221411884766539}{0.8455677672863762}{1.1}{pastel1}{62.21\%} & \databar{0.7993899411222246}{0.8455677672863762}{1.1}{pastel1}{79.94\%} & \databar{0.7212862680062135}{0.8455677672863762}{1.1}{pastel1}{72.13\%} & \databar{0.6113537191063729}{0.8455677672863762}{1.1}{pastel1}{61.14\%} & \databar{0.7766290488334031}{0.8455677672863762}{1.1}{pastel1}{77.66\%} & \databar{0.6722683318680828}{0.8455677672863762}{1.1}{pastel1}{67.23\%} & \databar{0.6002726868431524}{0.8455677672863762}{1.1}{pastel1}{60.03\%} & \databar{0.7282494009104636}{0.8455677672863762}{1.1}{pastel1}{72.82\%} & \databar{0.7486055212309402}{0.8455677672863762}{1.1}{pastel1}{74.86\%} & \databar{0.695033379741379}{0.8455677672863762}{1.1}{pastel1}{69.50\%} \\
StyEmb& \databar{0.6341624035205514}{0.8455677672863762}{1.1}{pastel1}{63.42\%} & \databar{0.6611845021579902}{0.8455677672863762}{1.1}{pastel1}{66.12\%} & \databar{0.5877768684548536}{0.8455677672863762}{1.1}{pastel1}{58.78\%} & \databar{0.6928292957309385}{0.8455677672863762}{1.1}{pastel1}{69.28\%} & \databar{0.699138890468227}{0.8455677672863762}{1.1}{pastel1}{69.91\%} & \databar{0.6390224264778208}{0.8455677672863762}{1.1}{pastel1}{63.90\%} & \databar{0.7017398651561395}{0.8455677672863762}{1.1}{pastel1}{70.17\%} & \databar{0.7102886556629218}{0.8455677672863762}{1.1}{pastel1}{71.03\%} & \databar{0.6495377646685632}{0.8455677672863762}{1.1}{pastel1}{64.95\%} & \databar{0.6609315149457526}{0.8455677672863762}{1.1}{pastel1}{66.09\%} & \databar{0.679768037878792}{0.8455677672863762}{1.1}{pastel1}{67.98\%} & \databar{0.6651254750111409}{0.8455677672863762}{1.1}{pastel1}{66.51\%} \\
StyDist & \databar{0.5991153790617431}{0.8455677672863762}{1.1}{pastel1}{59.91\%} & \databar{0.6377259464207259}{0.8455677672863762}{1.1}{pastel1}{63.77\%} & \databar{0.5594755704747407}{0.8455677672863762}{1.1}{pastel1}{55.95\%} & \databar{0.6485312277714168}{0.8455677672863762}{1.1}{pastel1}{64.85\%} & \databar{0.653306387713643}{0.8455677672863762}{1.1}{pastel1}{65.33\%} & \databar{0.619557831273375}{0.8455677672863762}{1.1}{pastel1}{61.96\%} & \databar{0.6899773198861169}{0.8455677672863762}{1.1}{pastel1}{69.00\%} & \databar{0.708061183774368}{0.8455677672863762}{1.1}{pastel1}{70.81\%} & \databar{0.6343904569654473}{0.8455677672863762}{1.1}{pastel1}{63.44\%} & \databar{0.6401528397954338}{0.8455677672863762}{1.1}{pastel1}{64.02\%} & \databar{0.6720209816145953}{0.8455677672863762}{1.1}{pastel1}{67.20\%} & \databar{0.6420286477046915}{0.8455677672863762}{1.1}{pastel1}{64.20\%} \\
MLangRob & \databar{0.6108498846777808}{0.8455677672863762}{1.1}{pastel1}{61.08\%} & \databar{0.6520787818391968}{0.8455677672863762}{1.1}{pastel1}{65.21\%} & \databar{0.5830001124118771}{0.8455677672863762}{1.1}{pastel1}{58.30\%} & \databar{0.8050298975237858}{0.8455677672863762}{1.1}{pastel1}{80.50\%} & \databar{0.6918118306356382}{0.8455677672863762}{1.1}{pastel1}{69.18\%} & \databar{0.5921487120412385}{0.8455677672863762}{1.1}{pastel1}{59.21\%} & \databar{0.8352215218781401}{0.8455677672863762}{1.1}{pastel1}{83.52\%} & \databar{0.7137185663983401}{0.8455677672863762}{1.1}{pastel1}{71.37\%} & \databar{0.573748649731444}{0.8455677672863762}{1.1}{pastel1}{57.37\%} & \databar{0.7265009240103549}{0.8455677672863762}{1.1}{pastel1}{72.65\%} & \databar{0.7433978586772669}{0.8455677672863762}{1.1}{pastel1}{74.34\%} & \databar{0.6843187945295512}{0.8455677672863762}{1.1}{pastel1}{68.43\%} \\
MLangLlama & \databar{0.5574983465608465}{0.8455677672863762}{1.1}{pastel1}{55.75\%} & \databar{0.5948700042307054}{0.8455677672863762}{1.1}{pastel1}{59.49\%} & \databar{0.5442378974045922}{0.8455677672863762}{1.1}{pastel1}{54.42\%} & \databar{0.8138219665501513}{0.8455677672863762}{1.1}{pastel1}{\textbf{81.38\%}} & \databar{0.6466007027285583}{0.8455677672863762}{1.1}{pastel1}{64.66\%} & \databar{0.5203992105284185}{0.8455677672863762}{1.1}{pastel1}{52.04\%} & \databar{0.816615342246257}{0.8455677672863762}{1.1}{pastel1}{81.66\%} & \databar{0.6968319016572979}{0.8455677672863762}{1.1}{pastel1}{69.68\%} & \databar{0.49755575065847235}{0.8455677672863762}{1.1}{pastel1}{49.76\%} & \databar{0.6531339512097183}{0.8455677672863762}{1.1}{pastel1}{65.31\%} & \databar{0.6855049969286914}{0.8455677672863762}{1.1}{pastel1}{68.55\%} & \databar{0.6388245518821554}{0.8455677672863762}{1.1}{pastel1}{63.88\%} \\
\midrule
PromptAV & \databar{0.5632697053027438}{0.8455677672863762}{1.1}{pastel1}{56.33\%} & \databar{0.5034695520185432}{0.8455677672863762}{1.1}{pastel1}{50.35\%} & \databar{0.5415631319570877}{0.8455677672863762}{1.1}{pastel1}{54.16\%} & \databar{0.6719178546959718}{0.8455677672863762}{1.1}{pastel1}{67.19\%} & \databar{0.624347547767468}{0.8455677672863762}{1.1}{pastel1}{62.43\%} & \databar{0.5422532781561451}{0.8455677672863762}{1.1}{pastel1}{54.23\%} & \databar{0.6155819852319486}{0.8455677672863762}{1.1}{pastel1}{61.56\%} & \databar{0.6201915075653692}{0.8455677672863762}{1.1}{pastel1}{62.02\%} & \databar{0.53613259345546}{0.8455677672863762}{1.1}{pastel1}{53.61\%} & \databar{0.5428885693090065}{0.8455677672863762}{1.1}{pastel1}{54.29\%} & \databar{0.5458097019484418}{0.8455677672863762}{1.1}{pastel1}{54.58\%} & \databar{0.5734023115825623}{0.8455677672863762}{1.1}{pastel1}{57.34\%} \\
LIP & \databar{0.46233048719236564}{0.8455677672863762}{1.1}{pastel1}{46.23\%} & \databar{0.2729836865375819}{0.8455677672863762}{1.1}{pastel1}{27.30\%} & \databar{0.3123755599800896}{0.8455677672863762}{1.1}{pastel1}{31.24\%} & \databar{0.11762870514820593}{0.8455677672863762}{1.1}{pastel1}{11.76\%} & \databar{0.3630031659882406}{0.8455677672863762}{1.1}{pastel1}{36.30\%} & \databar{0.34433811802232855}{0.8455677672863762}{1.1}{pastel1}{34.43\%} & \databar{0.09254498714652956}{0.8455677672863762}{1.1}{pastel1}{9.25\%} & \databar{0.3022337030846376}{0.8455677672863762}{1.1}{pastel1}{30.22\%} & \databar{0.3521854927150243}{0.8455677672863762}{1.1}{pastel1}{35.22\%} & \databar{0.3523555233742137}{0.8455677672863762}{1.1}{pastel1}{35.24\%} & \databar{0.33334444481482717}{0.8455677672863762}{1.1}{pastel1}{33.33\%} & \databar{0.30048398854582226}{0.8455677672863762}{1.1}{pastel1}{30.05\%} \\
\bottomrule
\end{tabular}

\end{adjustbox}
\end{table*}
\subsection{Effect of the number of anchor samples in nOSST-RS}\label{ssec:results-num-fewshot-samples}

\begin{figure}[!ht]
    \centering
    \includegraphics[width=\linewidth]{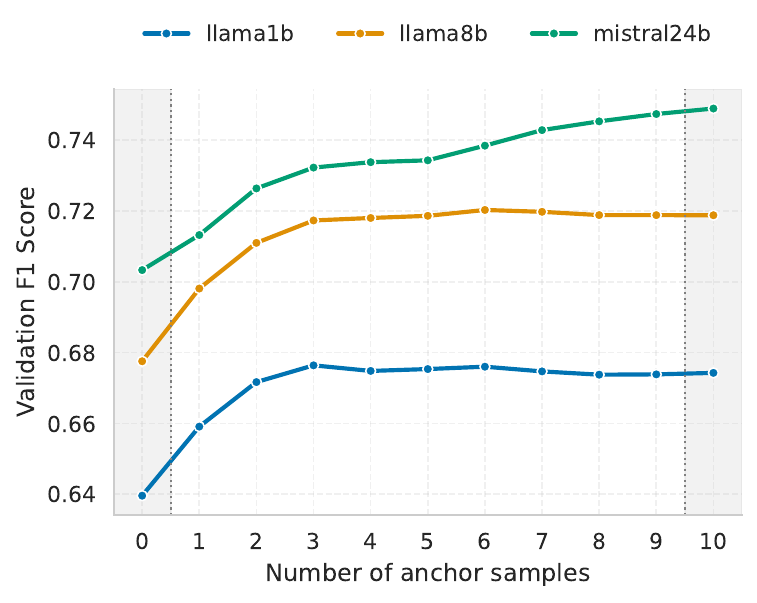}
    \caption{Average classification F1 score across all verification datasets by number of anchor examples in nOSST-RS. At $0$ we show the performance of nOSST-ZS, and at $10$ a single set of anchor samples was used.}
    \label{fig:results-num-fewshot-samples}
\end{figure}

\Cref{fig:results-num-fewshot-samples} presents the mean F1 classification score on all verification datasets as a function of the number of anchor examples provided to nOSST-RS. The point at zero corresponds to nOSST-ZS, and, for computational efficiency, we re-sample from a fixed pool of 10 examples. We observe that introducing a single anchor example yields a big improvement over nOSST-ZS with minimal additional cost. While F1 gains exhibit a pattern of diminishing returns up to a plateau, we observe that the Mistral24B base model shows an upward final trajectory. This may provide a guideline to choose the number of anchor examples for nOSST-RS in practice.

\subsection{AA multi-language performance}\label{sec:multilang}


\begin{table*}[!ht]
    \centering
    \caption{Accuracy and rank of correct candidate by language and method in the PAN18 and PAN19 AA datasets.}
    \begin{adjustbox}{width=\textwidth}
    \begin{tabular}{l|cc|cc|cc|cc|cc}
\toprule
 & \multicolumn{2}{c}{English} & \multicolumn{2}{|c}{French} & \multicolumn{2}{|c}{Italian} & \multicolumn{2}{|c}{Polish} & \multicolumn{2}{|c}{Spanish} \\
 & Accuracy & Mean rank & Accuracy & Mean rank & Accuracy & Mean rank & Accuracy & Mean rank & Accuracy & Mean rank \\
\midrule
OSST-Gemma1B & \databar{0.5818537502801625}{0.7974822726135836}{1.5}{pastel2}{58.19\%} & \databar{0.8718180573014961}{0.9608168096082128}{1.5}{pastel3}{0.87} & \databar{0.3626117113512071}{0.7974822726135836}{1.5}{pastel2}{36.26\%} & \databar{0.7620848339335734}{0.9608168096082128}{1.5}{pastel3}{0.76} & \databar{0.3068804500703235}{0.7974822726135836}{1.5}{pastel2}{30.69\%} & \databar{0.6881526801062666}{0.9608168096082128}{1.5}{pastel3}{0.69} & \databar{0.25}{0.7974822726135836}{1.5}{pastel2}{25.00\%} & \databar{0.6625740740740741}{0.9608168096082128}{1.5}{pastel3}{0.66} & \databar{0.42740358983472543}{0.7974822726135836}{1.5}{pastel2}{42.74\%} & \databar{0.7985187119657465}{0.9608168096082128}{1.5}{pastel3}{0.80} \\
OSST-Llama1B & \databar{0.6227818435182568}{0.7974822726135836}{1.5}{pastel2}{62.28\%} & \databar{0.8847999774787212}{0.9608168096082128}{1.5}{pastel3}{0.88} & \databar{0.3743497398959584}{0.7974822726135836}{1.5}{pastel2}{37.43\%} & \databar{0.7622028441006031}{0.9608168096082128}{1.5}{pastel3}{0.76} & \databar{0.41433473980309427}{0.7974822726135836}{1.5}{pastel2}{41.43\%} & \databar{0.8079440537583998}{0.9608168096082128}{1.5}{pastel3}{0.81} & \databar{0.3022222222222222}{0.7974822726135836}{1.5}{pastel2}{30.22\%} & \databar{0.725425925925926}{0.9608168096082128}{1.5}{pastel3}{0.73} & \databar{0.5331329348480974}{0.7974822726135836}{1.5}{pastel2}{53.31\%} & \databar{0.8461290567529217}{0.9608168096082128}{1.5}{pastel3}{0.85} \\
OSST-Gemma4B & \databar{0.7390241248371808}{0.7974822726135836}{1.5}{pastel2}{73.90\%} & \databar{0.9465091115535965}{0.9608168096082128}{1.5}{pastel3}{0.95} & \databar{0.5446178471388555}{0.7974822726135836}{1.5}{pastel2}{54.46\%} & \databar{0.872610896210336}{0.9608168096082128}{1.5}{pastel3}{\textbf{0.87}} & \databar{0.5826272855133615}{0.7974822726135836}{1.5}{pastel2}{58.26\%} & \databar{0.8652120643850602}{0.9608168096082128}{1.5}{pastel3}{0.87} & \databar{0.4055555555555556}{0.7974822726135836}{1.5}{pastel2}{40.56\%} & \databar{0.8035}{0.9608168096082128}{1.5}{pastel3}{0.80} & \databar{0.7127359233428837}{0.7974822726135836}{1.5}{pastel2}{71.27\%} & \databar{0.9251813186486819}{0.9608168096082128}{1.5}{pastel3}{0.93} \\
OSST-Llama8B & \databar{0.7974822726135836}{0.7974822726135836}{1.5}{pastel2}{\textbf{79.75\%}} & \databar{0.9608168096082128}{0.9608168096082128}{1.5}{pastel3}{\textbf{0.96}} & \databar{0.5523209283713485}{0.7974822726135836}{1.5}{pastel2}{\textbf{55.23\%}} & \databar{0.8572265943414403}{0.9608168096082128}{1.5}{pastel3}{0.86} & \databar{0.6166244725738397}{0.7974822726135836}{1.5}{pastel2}{\textbf{61.66\%}} & \databar{0.8924691358024691}{0.9608168096082128}{1.5}{pastel3}{\textbf{0.89}} & \databar{0.5188888888888888}{0.7974822726135836}{1.5}{pastel2}{\textbf{51.89\%}} & \databar{0.8751481481481482}{0.9608168096082128}{1.5}{pastel3}{\textbf{0.88}} & \databar{0.7439226667013427}{0.7974822726135836}{1.5}{pastel2}{\textbf{74.39\%}} & \databar{0.9391942304885995}{0.9608168096082128}{1.5}{pastel3}{\textbf{0.94}} \\
\midrule
STAR & \databar{0.693743361365829}{0.7980588390064218}{1.5}{pastel2}{69.37\%} & \databar{0.9304953544448668}{0.9609966098978855}{1.5}{pastel3}{0.93} & \databar{0.3859043617446979}{0.7980588390064218}{1.5}{pastel2}{38.59\%} & \databar{0.77255291005291}{0.9609966098978855}{1.5}{pastel3}{0.77} & \databar{0.24585654008438818}{0.7980588390064218}{1.5}{pastel2}{24.59\%} & \databar{0.7081683075480544}{0.9609966098978855}{1.5}{pastel3}{0.71} & \databar{0.42333333333333334}{0.7980588390064218}{1.5}{pastel2}{42.33\%} & \databar{0.8078888888888889}{0.9609966098978855}{1.5}{pastel3}{0.81} & \databar{0.35187055442617693}{0.7980588390064218}{1.5}{pastel2}{35.19\%} & \databar{0.798815272185066}{0.9609966098978855}{1.5}{pastel3}{0.80} \\
LUAR$_{CRUD}$ & \databar{0.5869941238951851}{0.7980588390064218}{1.5}{pastel2}{58.70\%} & \databar{0.8956105702755934}{0.9609966098978855}{1.5}{pastel3}{0.90} & \databar{0.2638555422168868}{0.7980588390064218}{1.5}{pastel2}{26.39\%} & \databar{0.7061180027566581}{0.9609966098978855}{1.5}{pastel3}{0.71} & \databar{0.23562869198312236}{0.7980588390064218}{1.5}{pastel2}{23.56\%} & \databar{0.6837955930614159}{0.9609966098978855}{1.5}{pastel3}{0.68} & \databar{0.2966666666666667}{0.7980588390064218}{1.5}{pastel2}{29.67\%} & \databar{0.7337222222222223}{0.9609966098978855}{1.5}{pastel3}{0.73} & \databar{0.21884901628472478}{0.7980588390064218}{1.5}{pastel2}{21.88\%} & \databar{0.651711068825634}{0.9609966098978855}{1.5}{pastel3}{0.65} \\
LUAR$_{MUD}$ & \databar{0.5828087391223847}{0.7980588390064218}{1.5}{pastel2}{58.28\%} & \databar{0.893650488159546}{0.9609966098978855}{1.5}{pastel3}{0.89} & \databar{0.32317927170868344}{0.7980588390064218}{1.5}{pastel2}{32.32\%} & \databar{0.7572317816015295}{0.9609966098978855}{1.5}{pastel3}{0.76} & \databar{0.27039662447257384}{0.7980588390064218}{1.5}{pastel2}{27.04\%} & \databar{0.6855541490857946}{0.9609966098978855}{1.5}{pastel3}{0.69} & \databar{0.26666666666666666}{0.7980588390064218}{1.5}{pastel2}{26.67\%} & \databar{0.7396111111111111}{0.9609966098978855}{1.5}{pastel3}{0.74} & \databar{0.26993329201107896}{0.7980588390064218}{1.5}{pastel2}{26.99\%} & \databar{0.7460686185411702}{0.9609966098978855}{1.5}{pastel3}{0.75} \\
StyEmb & \databar{0.5629513053138308}{0.7980588390064218}{1.5}{pastel2}{56.30\%} & \databar{0.8691878588652091}{0.9609966098978855}{1.5}{pastel3}{0.87} & \databar{0.2998199279711885}{0.7980588390064218}{1.5}{pastel2}{29.98\%} & \databar{0.7615907474100752}{0.9609966098978855}{1.5}{pastel3}{0.76} & \databar{0.28237974683544304}{0.7980588390064218}{1.5}{pastel2}{28.24\%} & \databar{0.7040450070323487}{0.9609966098978855}{1.5}{pastel3}{0.70} & \databar{0.26666666666666666}{0.7980588390064218}{1.5}{pastel2}{26.67\%} & \databar{0.7749444444444445}{0.9609966098978855}{1.5}{pastel3}{0.77} & \databar{0.2512125769915954}{0.7980588390064218}{1.5}{pastel2}{25.12\%} & \databar{0.728386534531775}{0.9609966098978855}{1.5}{pastel3}{0.73} \\
StyDist & \databar{0.5420341262339333}{0.7980588390064218}{1.5}{pastel2}{54.20\%} & \databar{0.8668256149229459}{0.9609966098978855}{1.5}{pastel3}{0.87} & \databar{0.4142156862745098}{0.7980588390064218}{1.5}{pastel2}{41.42\%} & \databar{0.7915749633186608}{0.9609966098978855}{1.5}{pastel3}{0.79} & \databar{0.34734177215189876}{0.7980588390064218}{1.5}{pastel2}{34.73\%} & \databar{0.750515236755743}{0.9609966098978855}{1.5}{pastel3}{0.75} & \databar{0.38333333333333336}{0.7980588390064218}{1.5}{pastel2}{38.33\%} & \databar{0.767611111111111}{0.9609966098978855}{1.5}{pastel3}{0.77} & \databar{0.35271794960707736}{0.7980588390064218}{1.5}{pastel2}{35.27\%} & \databar{0.7882721735072336}{0.9609966098978855}{1.5}{pastel3}{0.79} \\
MLangRob & \databar{0.47385474424814117}{0.7980588390064218}{1.5}{pastel2}{47.39\%} & \databar{0.8447243092301513}{0.9609966098978855}{1.5}{pastel3}{0.84} & \databar{0.3757002801120448}{0.7980588390064218}{1.5}{pastel2}{37.57\%} & \databar{0.8196845404828599}{0.9609966098978855}{1.5}{pastel3}{0.82} & \databar{0.3032573839662447}{0.7980588390064218}{1.5}{pastel2}{30.33\%} & \databar{0.7493380215658697}{0.9609966098978855}{1.5}{pastel3}{0.75} & \databar{0.23}{0.7980588390064218}{1.5}{pastel2}{23.00\%} & \databar{0.7761666666666667}{0.9609966098978855}{1.5}{pastel3}{0.78} & \databar{0.2802862505255584}{0.7980588390064218}{1.5}{pastel2}{28.03\%} & \databar{0.7790988016064628}{0.9609966098978855}{1.5}{pastel3}{0.78} \\
MLangLlama & \databar{0.44509057776824956}{0.7980588390064218}{1.5}{pastel2}{44.51\%} & \databar{0.8276253801819243}{0.9609966098978855}{1.5}{pastel3}{0.83} & \databar{0.2846138455382153}{0.7980588390064218}{1.5}{pastel2}{28.46\%} & \databar{0.7328442488106354}{0.9609966098978855}{1.5}{pastel3}{0.73} & \databar{0.1909704641350211}{0.7980588390064218}{1.5}{pastel2}{19.10\%} & \databar{0.6620951711204875}{0.9609966098978855}{1.5}{pastel3}{0.66} & \databar{0.19}{0.7980588390064218}{1.5}{pastel2}{19.00\%} & \databar{0.785}{0.9609966098978855}{1.5}{pastel3}{0.79} & \databar{0.23954192980707129}{0.7980588390064218}{1.5}{pastel2}{23.95\%} & \databar{0.7143420490540431}{0.9609966098978855}{1.5}{pastel3}{0.71} \\
\bottomrule
\end{tabular}

    \end{adjustbox}
    \label{tab:results-attribution-multilang}
\end{table*}

In \Cref{tab:results-attribution-multilang} we present results across various languages for AA, using the PAN18 and PAN19 multi-language AA datasets. The same contrastive models as in \cref{ssec:results-attribution} are included as baselines, and the results of OSST methods are averaged across three runs. As anticipated, performance declines for all methods outside of English. OSST-based methods with the larger backbones consistently outperform constrastive baselines in non-English settings, even for baselines trained on multiple languages. These results show OSST’s multilingual capabilities in AA, inherited from multilingual CLM pre-training, without requiring labeled authorship data in other languages.

\section{Advantages and disadvantages of OSST-based and contrastive methods}

OSST-based methods present several advantages:
\begin{enumerate*}[label=(\roman*)]
\item They are fully unsupervised with respect to authorship data, and can directly leverage any large open-source language model without fine-tuning.
\item In our limited experiments (verification datasets, up to 24B backbone models) they outperform prompting-based unsupervised baselines at similar model scales.
\item With larger backbones they are able to compete with or outperform contrastive models in most of our experimental settings.
\item They generalize across languages without language-specific authorship labeled data.
\end{enumerate*}

However, contrastive methods also have advantages. They are much faster in inference. Since they produce embeddings, they allow for vector retrieval, are easily applied to other authorship tasks (e.g. profiling and clustering) and are more interpretable. Finally, they do not have the limitation of requiring anchor samples as nOSST-RS. In fact, they might provide a solution to retrieve appropriate anchor samples, but we do not study this.

\section{Conclusions}\label{sec:conclusions}

We introduced OSST-based methods, a family of unsupervised approaches to authorship attribution and verification that take advantage of the scale and extensive CLM pre-training of LLMs.

In our experiments, OSST-based methods outperform LLM prompting alternatives of comparable scale in authorship verification. With sufficiently large backbone models, they are competitive with or surpass contrastive baselines in most of our experiments, and also present good results in authorship attribution in non-English languages. Our findings further suggest a scaling behavior: performance improves with model size and, for the verification random sampling variant (nOSST-RS), with the number of anchor examples, up to a saturation point. This behavior allows a practical trade-off between computational resources and stylistic accuracy.

\section*{Limitations}
We find the following limitations in our work.
\begin{enumerate*}[label=(\roman*)]
\item Model sizes were limited by hardware requirements. While small models are interesting in resource-constrained scenarios, further work needs to be conducted with larger models to properly observe scaling laws and push performance. We also restricted ourselves to non-commercial models due to economical constraints. Experiments with commercial models could be of interest due to the current gap with open-source ones.
\item Empirical comparisons are limited by lack of standardization and maturity of datasets and pre-processing in the field.
\item Contrastive baselines are trained for authorship with labeled data, and there may be intersection with the evaluation domains/distributions (perhaps even data), which may affect evaluation results.
\item We do not have larger contrastive models to compare performance at similar scales, although this is part of the advantage of unsupervised methods like OSST-based ones.
\item The random sampling (RS) variant for verification is studied with in-distribution anchors from the training set, which limits real-world applicability. Future research could study performance using a surrogate model to select anchors, such as already existing contrastive models.
\end{enumerate*}

\section*{Acknowledgments}
This work has been supported by the following projects: H2020 TMA-MSCA-DN TUAI project ”Towards an Understanding of Artificial Intelligence via a transparent, open and explainable perspective” (HORIZON-MSCA-2023-DN-01-01, Grant agreement nº: 101168344); by project PCI2022-134990-2 (MARTINI) of the CHISTERA IV Cofund 2021 program; by EMIF managed by the Calouste Gulbenkian Foundation, in the project MuseAI; and by Comunidad Autonoma de Madrid, CIRMA-CAM Project (TEC-2024/COM-404).

\printbibliography

\newpage
\appendix
\section{Decision boundary under distribution shifts in nOSST}\label{sec:cross-performance-AV}

\begin{figure}[!ht]
    \centering
    \includegraphics[width=\linewidth]{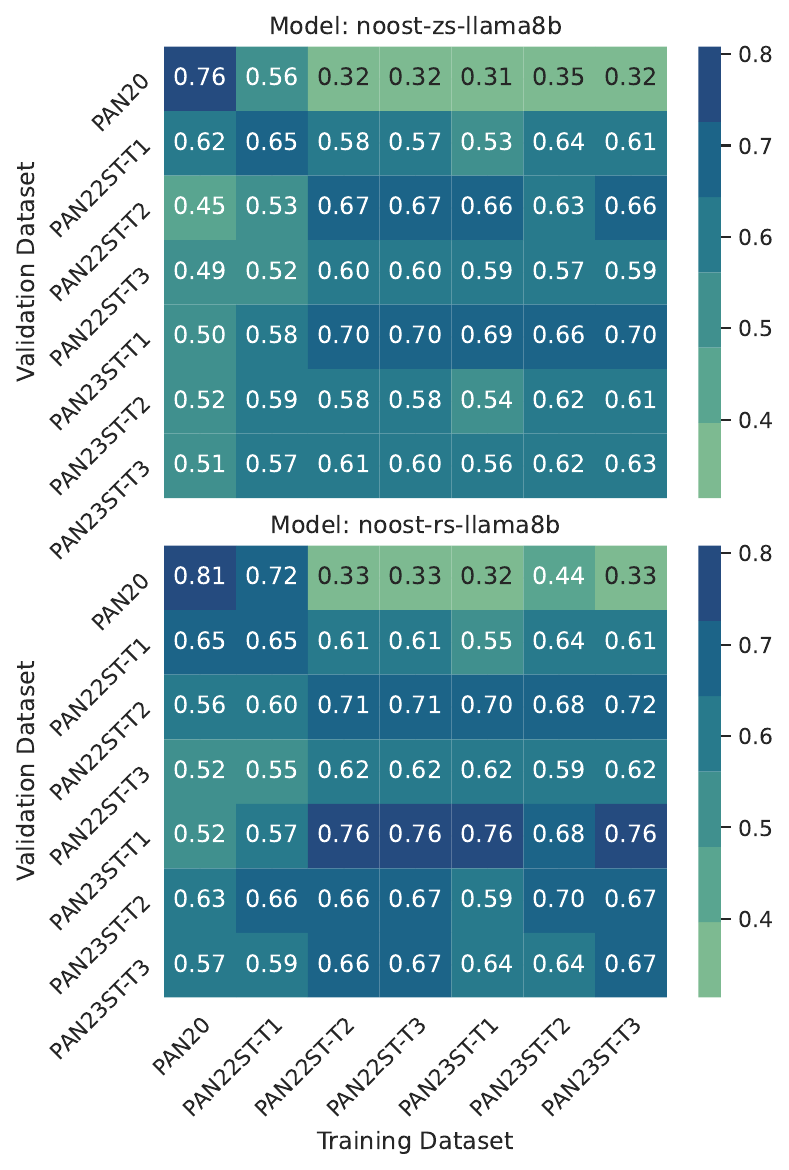}
    \caption{Cross-dataset macro-F1 of nOSST scores for authorship verification with a \texttt{LLaMA-8B} base model. The training set is used to select the decision boundary.}
    \label{fig:cross-performance-AV}
\end{figure}

\Cref{fig:cross-performance-AV} reports cross-dataset performance of nOSST scores for authorship verification, where the decision threshold is tuned on one dataset and evaluated on another. We consider three domains: FanFiction, StackExchange, and Reddit. As expected, thresholds selected in-distribution yield the best results. We observe the performance degradation is not too large in most pairs of datasets. However, this is not the case with the PAN20 dataset. This great difference could be partially explained by the great difference in text lengths (see appendix \Cref{fig:verification-dataset-lengths}), but not fully, as training in PAN22's first task does generalize to PAN20 in the RS variant.

\section{Model and prompt to obtain a neutrally styled version of a text}\label{sec:neutralize-prompt}

We use a \texttt{google/gemma-3-12b-it}~\cite{team2025Gemma3} model to obtain neutral versions of a given text. This decision was made for its multilingual capabilities and the inclusion of multilingual texts in some of the datasets used.

The prompt we used is the following:
\begin{lstlisting}[
    caption={Prompt to obtain a neutrally styled version of a text.},
    label={lst:neutralize-prompt},
]
Text: |{\color{cyan}[TEXT]}|
---
Paraphrase the previous text using the driest, simplest, and most neutral style possible.
Remove all stylistic features, emotion, or rhetorical devices. Replace as many words as possible with simpler or more common alternatives.
Preserve as much of the original meaning as possible. Write in |{\color{cyan}[LANGUAGE]}|. Output only the rewritten version.
\end{lstlisting}
\section{Ablation of the paraphrasing prompt}\label{sec:ablation-paraphrasing-prompt}

In this section we investigate the impact of our particular line of paraphrasing prompting in the performance of OSST-based methods. We compare against two baselines:

\begin{lstlisting}[
    caption={Version of the paraphrasing prompt asking to paraphrase the text.},
    label={lst:paraphrase-prompt},
]
Text: |{\color{cyan}[TEXT]}|
---
Paraphrase the previous text in a natural way, preserving the original meaning but changing the wording and structure. |{\color{cyan}[LANGUAGE]}|. Output only the rewritten version.
\end{lstlisting}

\begin{lstlisting}[
    caption={Version of the paraphrasing prompt asking to rephrase the text in the in the style of Shakespeare.},
    label={lst:shakespeare-prompt},
]
Text: |{\color{cyan}[TEXT]}|
---
Paraphrase the previous text in the style of Shakespeare, using archaic language and poetic devices. Preserve the original meaning but rewrite it in a way that evokes the tone and style of Shakespeare's works. Write in |{\color{cyan}[LANGUAGE]}|. Output only the rewritten version.
\end{lstlisting}

In \Cref{tab:paraphrasing-prompt-ablation-attribution} we show the results of our ablation for some AA datasets. Results are reported across 3 seeds. The neutralizing prompt seems to be the best performing one generally. 

\begin{table*}[!ht]
\caption{Authorship attribution results of OSST-based models with three different paraphrasing prompts: neutralizing, paraphrasing and rephrasing in the style of Shakespeare. We highlight in bold the best result across paraphrasing prompts for each dataset and base model.}
\label{tab:paraphrasing-prompt-ablation-attribution}
\centering

\begin{adjustbox}{width=\linewidth}
\begin{tabular}{ll|cc|cc|cc}
\toprule
 &  & \multicolumn{2}{c|}{PAN12} & \multicolumn{2}{c|}{PAN18-en} & \multicolumn{2}{c}{PAN19-en} \\
 Model & Paraph. prompt  & Accuracy & Mean rank & Accuracy & Mean rank & Accuracy & Mean rank \\
\midrule
\multirow[t]{3}{*}{OSST-Gemma1B} & neutral & \databar{0.3392857142857143}{0.8940936863543788}{2}{pastel2}{\textbf{33.93\%}\textsubscript{±1.03\%}} & \databar{0.6882440476190477}{0.9752206381534284}{2}{pastel3}{0.69\textsubscript{±0.03}} & \databar{0.47368421052631576}{0.8940936863543788}{2}{pastel2}{\textbf{47.37\%}\textsubscript{±1.66\%}} & \databar{0.84481658692185}{0.9752206381534284}{2}{pastel3}{\textbf{0.84}\textsubscript{±0.01}} & \databar{0.6934826883910387}{0.8940936863543788}{2}{pastel2}{\textbf{69.35\%}\textsubscript{±1.24\%}} & \databar{0.9016745870106359}{0.9752206381534284}{2}{pastel3}{\textbf{0.90}\textsubscript{±0.00}} \\
 & paraphrase & \databar{0.21428571428571427}{0.8940936863543788}{2}{pastel2}{21.43\%\textsubscript{±1.03\%}} & \databar{0.6748511904761906}{0.9752206381534284}{2}{pastel3}{0.67\textsubscript{±0.00}} & \databar{0.3492822966507177}{0.8940936863543788}{2}{pastel2}{34.93\%\textsubscript{±0.73\%}} & \databar{0.7586124401913876}{0.9752206381534284}{2}{pastel3}{0.76\textsubscript{±0.01}} & \databar{0.6323828920570265}{0.8940936863543788}{2}{pastel2}{63.24\%\textsubscript{±0.21\%}} & \databar{0.8716904276985743}{0.9752206381534284}{2}{pastel3}{0.87\textsubscript{±0.00}} \\
 & shakespeare & \databar{0.30357142857142855}{0.8940936863543788}{2}{pastel2}{30.36\%\textsubscript{±1.79\%}} & \databar{0.7258715986394558}{0.9752206381534284}{2}{pastel3}{\textbf{0.73}\textsubscript{±0.00}} & \databar{0.4688995215311005}{0.8940936863543788}{2}{pastel2}{46.89\%\textsubscript{±0.48\%}} & \databar{0.8070175438596491}{0.9752206381534284}{2}{pastel3}{0.81\textsubscript{±0.00}} & \databar{0.670061099796334}{0.8940936863543788}{2}{pastel2}{67.01\%\textsubscript{±0.26\%}} & \databar{0.8873048200950442}{0.9752206381534284}{2}{pastel3}{0.89\textsubscript{±0.00}} \\
\midrule
\multirow[t]{3}{*}{OSST-Llama1B} & neutral & \databar{0.39285714285714285}{0.8940936863543788}{2}{pastel2}{\textbf{39.29\%}\textsubscript{±8.25\%}} & \databar{0.7426658163265306}{0.9752206381534284}{2}{pastel3}{0.74\textsubscript{±0.04}} & \databar{0.5119617224880383}{0.8940936863543788}{2}{pastel2}{\textbf{51.20\%}\textsubscript{±2.09\%}} & \databar{0.8474481658692186}{0.9752206381534284}{2}{pastel3}{\textbf{0.85}\textsubscript{±0.02}} & \databar{0.7189409368635438}{0.8940936863543788}{2}{pastel2}{\textbf{71.89\%}\textsubscript{±0.88\%}} & \databar{0.9076714188730481}{0.9752206381534284}{2}{pastel3}{\textbf{0.91}\textsubscript{±0.01}} \\
 & paraphrase & \databar{0.30357142857142855}{0.8940936863543788}{2}{pastel2}{30.36\%\textsubscript{±0.00\%}} & \databar{0.7189625850340136}{0.9752206381534284}{2}{pastel3}{0.72\textsubscript{±0.00}} & \databar{0.40669856459330145}{0.8940936863543788}{2}{pastel2}{40.67\%\textsubscript{±1.44\%}} & \databar{0.8085326953748005}{0.9752206381534284}{2}{pastel3}{0.81\textsubscript{±0.01}} & \databar{0.6995926680244399}{0.8940936863543788}{2}{pastel2}{69.96\%\textsubscript{±0.12\%}} & \databar{0.8968092328581128}{0.9752206381534284}{2}{pastel3}{0.90\textsubscript{±0.00}} \\
 & shakespeare & \databar{0.39285714285714285}{0.8940936863543788}{2}{pastel2}{\textbf{39.29\%}\textsubscript{±0.00\%}} & \databar{0.7963435374149661}{0.9752206381534284}{2}{pastel3}{\textbf{0.80}\textsubscript{±0.00}} & \databar{0.42105263157894735}{0.8940936863543788}{2}{pastel2}{42.11\%\textsubscript{±0.28\%}} & \databar{0.8198564593301436}{0.9752206381534284}{2}{pastel3}{0.82\textsubscript{±0.00}} & \databar{0.6466395112016293}{0.8940936863543788}{2}{pastel2}{64.66\%\textsubscript{±0.06\%}} & \databar{0.8753111563702196}{0.9752206381534284}{2}{pastel3}{0.88\textsubscript{±0.00}} \\
\midrule
\multirow[t]{3}{*}{OSST-Gemma4B} & neutral & \databar{0.625}{0.8940936863543788}{2}{pastel2}{\textbf{62.50\%}\textsubscript{±3.09\%}} & \databar{0.8562925170068026}{0.9752206381534284}{2}{pastel3}{0.86\textsubscript{±0.02}} & \databar{0.6267942583732058}{0.8940936863543788}{2}{pastel2}{\textbf{62.68\%}\textsubscript{±1.54\%}} & \databar{0.9262360446570975}{0.9752206381534284}{2}{pastel3}{\textbf{0.93}\textsubscript{±0.01}} & \databar{0.8462321792260692}{0.8940936863543788}{2}{pastel2}{84.62\%\textsubscript{±0.24\%}} & \databar{0.9609640190088256}{0.9752206381534284}{2}{pastel3}{0.96\textsubscript{±0.00}} \\
 & paraphrase & \databar{0.5535714285714286}{0.8940936863543788}{2}{pastel2}{55.36\%\textsubscript{±0.00\%}} & \databar{0.8672406462585034}{0.9752206381534284}{2}{pastel3}{\textbf{0.87}\textsubscript{±0.00}} & \databar{0.6220095693779905}{0.8940936863543788}{2}{pastel2}{62.20\%\textsubscript{±3.19\%}} & \databar{0.9027910685805423}{0.9752206381534284}{2}{pastel3}{0.90\textsubscript{±0.00}} & \databar{0.8564154786150713}{0.8940936863543788}{2}{pastel2}{\textbf{85.64\%}\textsubscript{±0.16\%}} & \databar{0.9643584521384929}{0.9752206381534284}{2}{pastel3}{\textbf{0.96}\textsubscript{±0.00}} \\
 & shakespeare & \databar{0.5}{0.8940936863543788}{2}{pastel2}{50.00\%\textsubscript{±0.00\%}} & \databar{0.8551232993197279}{0.9752206381534284}{2}{pastel3}{0.86\textsubscript{±0.00}} & \databar{0.6267942583732058}{0.8940936863543788}{2}{pastel2}{\textbf{62.68\%}\textsubscript{±0.73\%}} & \databar{0.9129186602870814}{0.9752206381534284}{2}{pastel3}{0.91\textsubscript{±0.00}} & \databar{0.7505091649694501}{0.8940936863543788}{2}{pastel2}{75.05\%\textsubscript{±0.46\%}} & \databar{0.9229463679565513}{0.9752206381534284}{2}{pastel3}{0.92\textsubscript{±0.00}} \\
\midrule
\multirow[t]{3}{*}{OSST-Llama8B} & neutral & \databar{0.6964285714285714}{0.8940936863543788}{2}{pastel2}{69.64\%\textsubscript{±3.09\%}} & \databar{0.9349489795918368}{0.9752206381534284}{2}{pastel3}{\textbf{0.93}\textsubscript{±0.01}} & \databar{0.6985645933014354}{0.8940936863543788}{2}{pastel2}{69.86\%\textsubscript{±1.10\%}} & \databar{0.9456937799043061}{0.9752206381534284}{2}{pastel3}{0.95\textsubscript{±0.00}} & \databar{0.8940936863543788}{0.8940936863543788}{2}{pastel2}{\textbf{89.41\%}\textsubscript{±0.71\%}} & \databar{0.9752206381534284}{0.9752206381534284}{2}{pastel3}{\textbf{0.98}\textsubscript{±0.00}} \\
 & paraphrase & \databar{0.5892857142857143}{0.8940936863543788}{2}{pastel2}{58.93\%\textsubscript{±0.00\%}} & \databar{0.9076318027210883}{0.9752206381534284}{2}{pastel3}{0.91\textsubscript{±0.00}} & \databar{0.7416267942583732}{0.8940936863543788}{2}{pastel2}{74.16\%\textsubscript{±1.10\%}} & \databar{0.9503189792663477}{0.9752206381534284}{2}{pastel3}{0.95\textsubscript{±0.00}} & \databar{0.8747454175152749}{0.8940936863543788}{2}{pastel2}{87.47\%\textsubscript{±0.71\%}} & \databar{0.9663951120162932}{0.9752206381534284}{2}{pastel3}{0.97\textsubscript{±0.00}} \\
 & shakespeare & \databar{0.7142857142857143}{0.8940936863543788}{2}{pastel2}{\textbf{71.43\%}\textsubscript{±0.00\%}} & \databar{0.9217687074829932}{0.9752206381534284}{2}{pastel3}{0.92\textsubscript{±0.00}} & \databar{0.784688995215311}{0.8940936863543788}{2}{pastel2}{\textbf{78.47\%}\textsubscript{±0.00\%}} & \databar{0.9579744816586921}{0.9752206381534284}{2}{pastel3}{\textbf{0.96}\textsubscript{±0.00}} & \databar{0.8625254582484725}{0.8940936863543788}{2}{pastel2}{86.25\%\textsubscript{±0.00\%}} & \databar{0.9687712152070604}{0.9752206381534284}{2}{pastel3}{0.97\textsubscript{±0.00}} \\
\bottomrule
\end{tabular}
\end{adjustbox}
\end{table*}

In \Cref{tab:paraphrasing-prompt-ablation-verification} we show some results in AV datasets. In this case the neutral and paraphrasing prompts are very competitive on average, with superiority varying between datasets and models. The Shakespeare strategy seems to lag slightly behind.

\begin{table*}[!ht]
\caption{Authorship verification results (F1 score) of OSST-based models with three different prompt versions: neutralizing (N), paraphrasing (P) and rephrasing in the style of Shakespeare (S). Base models are Llama1B and Llama8B. We highlight in bold the best result across paraphrasing prompts for each dataset and base model.}
\label{tab:paraphrasing-prompt-ablation-verification}
\centering

\begin{adjustbox}{width=\linewidth}
\begin{tabular}{ll|ccc|ccc|ccc|cc|c}
\toprule
 Model & Prompt  & 22ST-T1 & 22ST-T2 & 22ST-T3 & 23ST-T1 & 23ST-T2 & 23ST-T3 & 24ST-T1 & 24ST-T2 & 24ST-T3 & 20AV & 21AV & Avg \\
\midrule
\multirow[t]{3}{*}{ZS-1B} & N & \databar{0.6226331730419634}{0.8306789136772068}{1.5}{pastel1}{\textbf{62.26\%}} & \databar{0.6257111777048504}{0.8306789136772068}{1.5}{pastel1}{\textbf{62.57\%}} & \databar{0.5830122673300138}{0.8306789136772068}{1.5}{pastel1}{58.30\%} & \databar{0.6900613770375695}{0.8306789136772068}{1.5}{pastel1}{69.01\%} & \databar{0.5582210180460465}{0.8306789136772068}{1.5}{pastel1}{55.82\%} & \databar{0.5896835207515678}{0.8306789136772068}{1.5}{pastel1}{\textbf{58.97\%}} & \databar{0.7320525793725154}{0.8306789136772068}{1.5}{pastel1}{73.21\%} & \databar{0.6712742312494779}{0.8306789136772068}{1.5}{pastel1}{67.13\%} & \databar{0.6126927172805838}{0.8306789136772068}{1.5}{pastel1}{\textbf{61.27\%}} & \databar{0.6764963913709701}{0.8306789136772068}{1.5}{pastel1}{\textbf{67.65\%}} & \databar{0.6733727946355741}{0.8306789136772068}{1.5}{pastel1}{\textbf{67.34\%}} & \databar{0.6395646588928302}{0.8306789136772068}{1.5}{pastel1}{\textbf{63.96\%}} \\
 & P & \databar{0.6088951764570963}{0.8306789136772068}{1.5}{pastel1}{60.89\%} & \databar{0.6113139066641575}{0.8306789136772068}{1.5}{pastel1}{61.13\%} & \databar{0.5830563139601602}{0.8306789136772068}{1.5}{pastel1}{\textbf{58.31\%}} & \databar{0.7077548189211493}{0.8306789136772068}{1.5}{pastel1}{70.78\%} & \databar{0.608199769459934}{0.8306789136772068}{1.5}{pastel1}{\textbf{60.82\%}} & \databar{0.5737537441561433}{0.8306789136772068}{1.5}{pastel1}{57.38\%} & \databar{0.7423566620712027}{0.8306789136772068}{1.5}{pastel1}{74.24\%} & \databar{0.6635931701935613}{0.8306789136772068}{1.5}{pastel1}{66.36\%} & \databar{0.6000481016406112}{0.8306789136772068}{1.5}{pastel1}{60.00\%} & \databar{0.6595537158625939}{0.8306789136772068}{1.5}{pastel1}{65.96\%} & \databar{0.6547114175801123}{0.8306789136772068}{1.5}{pastel1}{65.47\%} & \databar{0.6375669815424293}{0.8306789136772068}{1.5}{pastel1}{63.76\%} \\
 & S & \databar{0.6175169588759838}{0.8306789136772068}{1.5}{pastel1}{61.75\%} & \databar{0.596892899114091}{0.8306789136772068}{1.5}{pastel1}{59.69\%} & \databar{0.5623779328468896}{0.8306789136772068}{1.5}{pastel1}{56.24\%} & \databar{0.7161380392636412}{0.8306789136772068}{1.5}{pastel1}{\textbf{71.61\%}} & \databar{0.5431372725192459}{0.8306789136772068}{1.5}{pastel1}{54.31\%} & \databar{0.5357219918627295}{0.8306789136772068}{1.5}{pastel1}{53.57\%} & \databar{0.7512729102881348}{0.8306789136772068}{1.5}{pastel1}{\textbf{75.13\%}} & \databar{0.7375113766370347}{0.8306789136772068}{1.5}{pastel1}{\textbf{73.75\%}} & \databar{0.5604699403268534}{0.8306789136772068}{1.5}{pastel1}{56.05\%} & \databar{0.6362362128033896}{0.8306789136772068}{1.5}{pastel1}{63.62\%} & \databar{0.6459375860716031}{0.8306789136772068}{1.5}{pastel1}{64.59\%} & \databar{0.627564829146327}{0.8306789136772068}{1.5}{pastel1}{62.76\%} \\
\midrule
\multirow[t]{3}{*}{ZS-8B} & N & \databar{0.6457309693608932}{0.8306789136772068}{1.5}{pastel1}{\textbf{64.57\%}} & \databar{0.6687141004777233}{0.8306789136772068}{1.5}{pastel1}{66.87\%} & \databar{0.5998814102036238}{0.8306789136772068}{1.5}{pastel1}{\textbf{59.99\%}} & \databar{0.6863386423808745}{0.8306789136772068}{1.5}{pastel1}{68.63\%} & \databar{0.6153526176950219}{0.8306789136772068}{1.5}{pastel1}{61.54\%} & \databar{0.6292736136588788}{0.8306789136772068}{1.5}{pastel1}{62.93\%} & \databar{0.7541517665290601}{0.8306789136772068}{1.5}{pastel1}{\textbf{75.42\%}} & \databar{0.6927772177962457}{0.8306789136772068}{1.5}{pastel1}{69.28\%} & \databar{0.6335776743502647}{0.8306789136772068}{1.5}{pastel1}{63.36\%} & \databar{0.7621592388972664}{0.8306789136772068}{1.5}{pastel1}{76.22\%} & \databar{0.7650142117987493}{0.8306789136772068}{1.5}{pastel1}{76.50\%} & \databar{0.6775428602862366}{0.8306789136772068}{1.5}{pastel1}{67.75\%} \\
 & P & \databar{0.6280611354390078}{0.8306789136772068}{1.5}{pastel1}{62.81\%} & \databar{0.6732493860846787}{0.8306789136772068}{1.5}{pastel1}{\textbf{67.32\%}} & \databar{0.5991542690202876}{0.8306789136772068}{1.5}{pastel1}{59.92\%} & \databar{0.7151923749296931}{0.8306789136772068}{1.5}{pastel1}{\textbf{71.52\%}} & \databar{0.6549093052731703}{0.8306789136772068}{1.5}{pastel1}{\textbf{65.49\%}} & \databar{0.6352650307481044}{0.8306789136772068}{1.5}{pastel1}{\textbf{63.53\%}} & \databar{0.749389397378025}{0.8306789136772068}{1.5}{pastel1}{74.94\%} & \databar{0.7333429537753837}{0.8306789136772068}{1.5}{pastel1}{73.33\%} & \databar{0.6407691707935571}{0.8306789136772068}{1.5}{pastel1}{\textbf{64.08\%}} & \databar{0.7719655823513327}{0.8306789136772068}{1.5}{pastel1}{\textbf{77.20\%}} & \databar{0.7818720084165408}{0.8306789136772068}{1.5}{pastel1}{\textbf{78.19\%}} & \databar{0.6893791467463437}{0.8306789136772068}{1.5}{pastel1}{\textbf{68.94\%}} \\
 & S & \databar{0.6356917026506894}{0.8306789136772068}{1.5}{pastel1}{63.57\%} & \databar{0.6388019130068079}{0.8306789136772068}{1.5}{pastel1}{63.88\%} & \databar{0.5829074197718944}{0.8306789136772068}{1.5}{pastel1}{58.29\%} & \databar{0.6857320467388938}{0.8306789136772068}{1.5}{pastel1}{68.57\%} & \databar{0.5759255407994498}{0.8306789136772068}{1.5}{pastel1}{57.59\%} & \databar{0.5791579024628619}{0.8306789136772068}{1.5}{pastel1}{57.92\%} & \databar{0.7122702039885174}{0.8306789136772068}{1.5}{pastel1}{71.23\%} & \databar{0.7587135226453343}{0.8306789136772068}{1.5}{pastel1}{\textbf{75.87\%}} & \databar{0.6050726681220162}{0.8306789136772068}{1.5}{pastel1}{60.51\%} & \databar{0.7574579280394851}{0.8306789136772068}{1.5}{pastel1}{75.75\%} & \databar{0.7661368752375292}{0.8306789136772068}{1.5}{pastel1}{76.61\%} & \databar{0.6634425203148617}{0.8306789136772068}{1.5}{pastel1}{66.34\%} \\
\bottomrule
\end{tabular}
\end{adjustbox}
\end{table*}

\section{Measuring semantic similarity with different paraphrasing prompts}

In \Cref{tab:paraphrasing-prompt-similarity-attribution,tab:paraphrasing-prompt-similarity-verification} we show semantic similarity statistics in AA and AV sets for paraphrases with three different prompt versions: neutralizing, paraphrasing and rephrasing in the style of Shakespeare. This is measured with a \texttt{sentence-transformers/all-mpnet-base-v2} model for semantic similarity.

\begin{table*}[!ht]
\caption{Semantic similarity statistics in authorship attributions sets for paraphrases with three different prompt versions: neutralizing, paraphrasing and rephrasing in the style of Shakespeare. Measured with a \texttt{sentence-transformers/all-mpnet-base-v2} model for semantic similarity.}
\label{tab:paraphrasing-prompt-similarity-attribution}
\centering

\begin{adjustbox}{width=0.7\linewidth}
\begin{tabular}{ll|cccccc}
\toprule
 &  & mean±std & min & q25 & median & q75 & max \\
dataset & variant &  &  &  &  &  &  \\
\midrule
\multirow[t]{3}{*}{11AA} & neutral & 0.700 ± 0.158 & 0.002 & 0.604 & 0.724 & 0.818 & 1.000 \\
 & paraphrase & 0.683 ± 0.158 & -0.017 & 0.585 & 0.700 & 0.805 & 1.000 \\
 & shakespeare & 0.451 ± 0.151 & -0.023 & 0.346 & 0.450 & 0.558 & 0.904 \\
\midrule
\multirow[t]{3}{*}{12AA} & neutral & 0.790 ± 0.076 & 0.470 & 0.743 & 0.813 & 0.840 & 0.938 \\
 & paraphrase & 0.877 ± 0.065 & 0.639 & 0.848 & 0.896 & 0.923 & 0.960 \\
 & shakespeare & 0.753 ± 0.103 & 0.446 & 0.692 & 0.768 & 0.825 & 0.924 \\
\midrule
\multirow[t]{3}{*}{18AA} & neutral & 0.820 ± 0.097 & 0.309 & 0.776 & 0.839 & 0.890 & 0.970 \\
 & paraphrase & 0.783 ± 0.165 & 0.235 & 0.694 & 0.849 & 0.906 & 0.973 \\
 & shakespeare & 0.721 ± 0.166 & 0.232 & 0.603 & 0.766 & 0.855 & 0.971 \\
\midrule
\multirow[t]{3}{*}{19AA} & neutral & 0.794 ± 0.092 & 0.309 & 0.747 & 0.812 & 0.859 & 0.980 \\
 & paraphrase & 0.748 ± 0.182 & 0.169 & 0.598 & 0.823 & 0.890 & 0.987 \\
 & shakespeare & 0.661 ± 0.173 & 0.130 & 0.524 & 0.716 & 0.802 & 1.000 \\
\bottomrule
\end{tabular}
\end{adjustbox}
\end{table*}

\begin{table*}[!ht]
\caption{Semantic similarity statistics in authorship verification test sets for paraphrases with three different prompt versions: neutralizing, paraphrasing and rephrasing in the style of Shakespeare. Measured with a \texttt{sentence-transformers/all-mpnet-base-v2} model for semantic similarity.}
\label{tab:paraphrasing-prompt-similarity-verification}
\centering

\begin{adjustbox}{width=0.7\linewidth}
\begin{tabular}{ll|cccccc}
\toprule
 &  & mean±std & min & q25 & median & q75 & max \\
dataset & variant &  &  &  &  &  &  \\
\midrule
\multirow[t]{3}{*}{20AV} & neutral & 0.786 ± 0.093 & 0.070 & 0.739 & 0.803 & 0.852 & 0.966 \\
 & paraphrase & 0.883 ± 0.071 & 0.178 & 0.852 & 0.902 & 0.932 & 0.984 \\
 & shakespeare & 0.754 ± 0.084 & 0.221 & 0.706 & 0.767 & 0.815 & 0.941 \\
\midrule
\multirow[t]{3}{*}{21AV} & neutral & 0.781 ± 0.096 & 0.021 & 0.733 & 0.799 & 0.849 & 0.969 \\
 & paraphrase & 0.880 ± 0.072 & 0.004 & 0.850 & 0.900 & 0.931 & 0.987 \\
 & shakespeare & 0.742 ± 0.090 & 0.052 & 0.692 & 0.755 & 0.807 & 0.948 \\
\midrule
\multirow[t]{3}{*}{22SC-T1} & neutral & 0.764 ± 0.139 & 0.079 & 0.697 & 0.792 & 0.865 & 0.994 \\
 & paraphrase & 0.770 ± 0.131 & 0.084 & 0.708 & 0.796 & 0.867 & 0.981 \\
 & shakespeare & 0.486 ± 0.170 & -0.059 & 0.371 & 0.495 & 0.609 & 0.929 \\
\midrule
\multirow[t]{3}{*}{22SC-T2} & neutral & 0.780 ± 0.118 & 0.042 & 0.715 & 0.801 & 0.867 & 0.994 \\
 & paraphrase & 0.795 ± 0.113 & 0.052 & 0.739 & 0.817 & 0.876 & 0.993 \\
 & shakespeare & 0.499 ± 0.162 & -0.010 & 0.388 & 0.511 & 0.618 & 0.930 \\
\midrule
\multirow[t]{3}{*}{22SC-T3} & neutral & 0.751 ± 0.142 & -0.018 & 0.672 & 0.776 & 0.857 & 1.000 \\
 & paraphrase & 0.750 ± 0.142 & 0.015 & 0.675 & 0.777 & 0.856 & 0.996 \\
 & shakespeare & 0.452 ± 0.175 & -0.072 & 0.322 & 0.454 & 0.582 & 0.949 \\
\midrule
\multirow[t]{3}{*}{23SC-T1} & neutral & 0.755 ± 0.124 & 0.133 & 0.691 & 0.781 & 0.835 & 0.984 \\
 & paraphrase & 0.771 ± 0.118 & 0.117 & 0.699 & 0.796 & 0.854 & 0.976 \\
 & shakespeare & 0.521 ± 0.155 & 0.045 & 0.398 & 0.523 & 0.637 & 0.898 \\
\midrule
\multirow[t]{3}{*}{23SC-T2} & neutral & 0.762 ± 0.120 & 0.031 & 0.694 & 0.784 & 0.850 & 0.989 \\
 & paraphrase & 0.774 ± 0.120 & 0.209 & 0.704 & 0.796 & 0.867 & 0.988 \\
 & shakespeare & 0.510 ± 0.168 & 0.012 & 0.384 & 0.515 & 0.641 & 0.907 \\
\midrule
\multirow[t]{3}{*}{23SC-T3} & neutral & 0.768 ± 0.107 & 0.171 & 0.706 & 0.785 & 0.848 & 0.977 \\
 & paraphrase & 0.775 ± 0.110 & 0.185 & 0.708 & 0.792 & 0.858 & 0.977 \\
 & shakespeare & 0.513 ± 0.150 & -0.009 & 0.406 & 0.523 & 0.626 & 0.918 \\
\midrule
\multirow[t]{3}{*}{24SC-T1} & neutral & 0.735 ± 0.123 & 0.165 & 0.656 & 0.770 & 0.832 & 0.969 \\
 & paraphrase & 0.749 ± 0.116 & 0.186 & 0.685 & 0.773 & 0.826 & 0.974 \\
 & shakespeare & 0.496 ± 0.141 & 0.107 & 0.386 & 0.493 & 0.605 & 0.870 \\
\midrule
\multirow[t]{3}{*}{24SC-T2} & neutral & 0.714 ± 0.120 & 0.167 & 0.628 & 0.736 & 0.801 & 0.974 \\
 & paraphrase & 0.739 ± 0.107 & 0.164 & 0.685 & 0.745 & 0.816 & 0.980 \\
 & shakespeare & 0.492 ± 0.137 & 0.053 & 0.401 & 0.507 & 0.581 & 0.871 \\
\midrule
\multirow[t]{3}{*}{24SC-T3} & neutral & 0.755 ± 0.111 & 0.214 & 0.691 & 0.773 & 0.838 & 0.980 \\
 & paraphrase & 0.754 ± 0.112 & 0.108 & 0.681 & 0.770 & 0.839 & 0.980 \\
 & shakespeare & 0.527 ± 0.138 & 0.078 & 0.433 & 0.536 & 0.633 & 0.883 \\
\bottomrule
\end{tabular}
\end{adjustbox}
\end{table*}

\section{Datasets}\label{sec:datasets-used-appendix}

In this section we list and briefly describe the datasets that we used in this paper for evaluation.

\paragraph{Authorship attribution datasets.}
The PAN-AA-2011~\cite{argamon_2011_3713246} and PAN-AA-2012~\cite{juola_2012_3713273} draw from corporate emails and fiction domain, respectively. We also use the PAN-AA-2018 dataset~\cite{kestemont_2018_3737849,kestemont2018overview}, which established a challenging cross-fandom scenario within FanFiction, where all unknown documents originate from a single fandom, while candidate authors’ documents span non-overlapping fandoms. This design amplifies complexity by introducing domain shift and reducing stylistic overlap, as authors often emulate the source material. The PAN-AA-2019 task~\cite{kestemont_2019_3530313,kestemont2019overview} further advanced these challenges through an open-set evaluation.

\paragraph{Authorship verification datasets.}
The PAN-AV-2020~\cite{bischoff2020ImportanceSuppressing} and PAN-AV-2021~\cite{bevendorff2020PAN20Authorship} authorship verification tasks centered on FanFiction, with the latter introducing an out-of-distribution test set featuring unseen authors and novel topics.

We also consider style-change detection tasks. These involve determining whether consecutive paragraphs in a text are authored by the same person. We transform these datasets to standard authorship verification format by constructing independent examples for each pair of consecutive paragraphs.

We consider data from the PAN-SC-2022~\cite{zangerle2022PAN22Authorship}, PAN-SC-2023~\cite{zangerle2023PAN23MultiAuthor}, and PAN-SC-2024~\cite{zangerle2024PAN24MultiAuthor} tasks. The 2022 dataset features StackExchange posts, while the 2023 and 2024 datasets focus on Reddit. The 2023 and 2024 datasets provide three subtasks, curating them to control for topical cues. The first tasks don't account for this bias, while the second and third tasks progressively increase in difficulty in this regard.
\section{Data preprocessing}\label{sec:data_preprocessing}

\paragraph{Processing FanFiction data.} To preprocess the FanFiction data used in the PAN18 and PAN19 authorship attribution tasks, we first remove all lines that do not contain alphabetic characters, thereby eliminating chapter separators that should not inform authorship prediction. Next, we remove line breaks to normalize paragraph formatting across authors, as the number of line breaks can vary. Finally, we truncate text at the paragraph level to a maximum of 384 tokens.

\paragraph{Enron emails:} To preprocess the Enron emails data used in the PAN11 authorship attribution tasks, we remove surrounding whitespace, truncate to 384 tokens, and remove author texts under 20 characters, and select a maximum of 50 texts per author to avoid massive computational costs.

\paragraph{Cleaning style change data.} As some paragraphs in the style change data where purely random symbols, we remove any sample where the paragraph does not contain at least one alphabetic character.

\section{Summary of candidates and number of texts in attribution datasets}

In \Cref{tab:summary-attribution-datasets} we show the summary of the datasets used for authorship attribution, including the number of targets to predict for each amount of candidates and texts per candidate. \emph{For problems with more candidates, the accuracy value will be naturally lower, but the mean rank value should be more stable across settings}.

\begin{table}[!ht]
\caption{Summary of the datasets used for authorship attribution, including the number of targets to predict for each amount of candidates and texts per candidate.}
\label{tab:summary-attribution-datasets}
\centering

\begin{adjustbox}{width=\linewidth}
\begin{tabular}{cccc}
\toprule
Dataset                   & Number of texts              & \# candidates & \# targets \\
\midrule
\multirow{2}{*}{PAN11-en} & \multirow{2}{*}{Varies 1-50} & 26            & 495        \\
                          &                              & 72            & 1300       \\
\midrule
\multirow{3}{*}{PAN12-en} & \multirow{3}{*}{2}           & 3             & 16         \\
                          &                              & 8             & 25         \\
                          &                              & 14            & 30         \\
\midrule
\multirow{4}{*}{PAN18-en} & \multirow{4}{*}{7}           & 5             & 16         \\
                          &                              & 10            & 40         \\
                          &                              & 15            & 74         \\
                          &                              & 20            & 79         \\
\midrule
\multirow{4}{*}{PAN18-fr} & \multirow{4}{*}{7}           & 5             & 19         \\
                          &                              & 10            & 47         \\
                          &                              & 15            & 77         \\
                          &                              & 20            & 95         \\
\midrule
\multirow{4}{*}{PAN18-it} & \multirow{4}{*}{7}           & 5             & 40         \\
                          &                              & 10            & 30         \\
                          &                              & 15            & 72         \\
                          &                              & 20            & 95         \\
\midrule
\multirow{4}{*}{PAN18-pl} & \multirow{4}{*}{7}           & 5             & 33         \\
                          &                              & 10            & 42         \\
                          &                              & 15            & 100        \\
                          &                              & 20            & 125        \\
\midrule
\multirow{4}{*}{PAN18-sp} & \multirow{4}{*}{7}           & 5             & 25         \\
                          &                              & 10            & 70         \\
                          &                              & 15            & 103        \\
                          &                              & 20            & 133        \\
\midrule
PAN19-en                  & 7                            & 9             & 1446       \\
PAN19-fr                  & 7                            & 9             & 920        \\
PAN19-it                  & 7                            & 9             & 551        \\
PAN19-sp                  & 7                            & 9             & 1134      \\
\bottomrule
\end{tabular}
\end{adjustbox}
\end{table}

\section{Verification dataset lengths}\label{sec:verification-dataset-lengths}

\begin{figure*}[!ht]
    \centering
    \includegraphics[width=\linewidth]{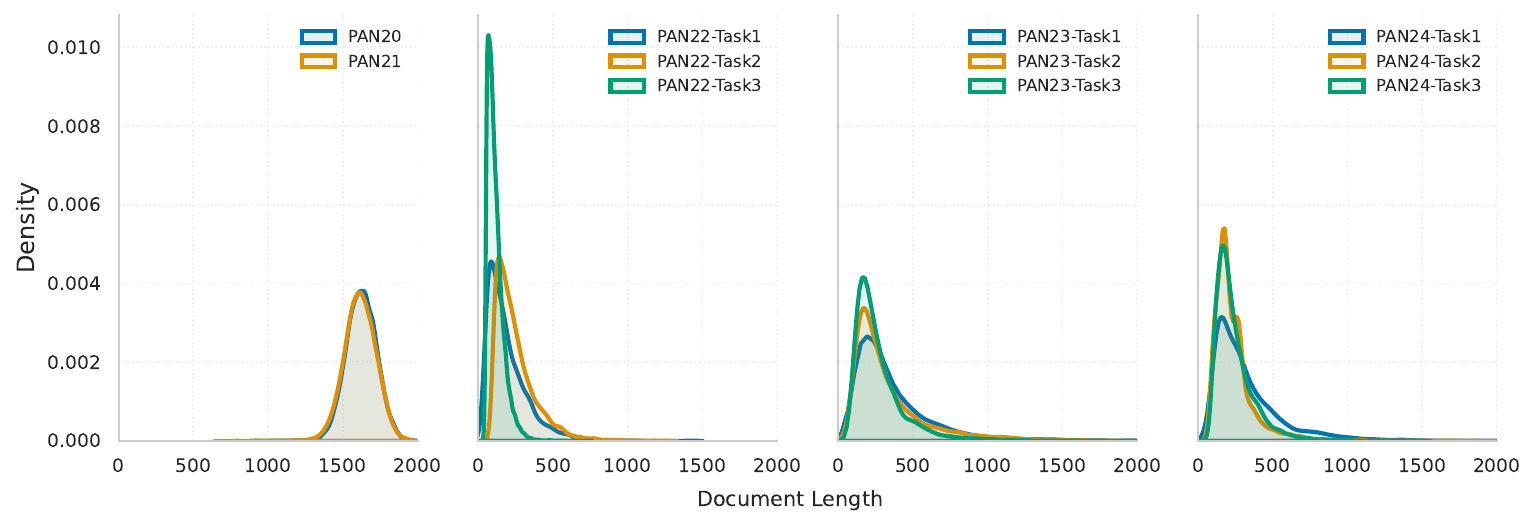}
    \caption{Length distributions (in characters) of texts in different verification datasets after truncation.}
    \label{fig:verification-dataset-lengths}
\end{figure*}

\Cref{fig:verification-dataset-lengths} presents the length distributions (in characters) of texts in different verification datasets after truncation.

\section{Base models used in OSST}\label{sec:base-models}
The following models were used as base for the OSST-based methods in this work:

\begin{itemize}
    \item \texttt{google/gemma-3-1b-pt}~\cite{team2025Gemma3}
    \item \texttt{google/gemma-3-4b-pt}~\cite{team2025Gemma3}
    \item \texttt{meta-llama/Llama-3.2-1B}~\cite{grattafiori2024Llama3}
    \item \texttt{meta-llama/Llama-3.2-8B}~\cite{grattafiori2024Llama3}
    \item \texttt{mistralai/Mistral-Small-24B-Base-2501}
\end{itemize}

With this choice we cover several model sizes and more than one model family. Llama, Gemma and Mistral models are some of the most well known open-source families. We also wanted to focus on models trained primarily on English data, since they will likely perform better on the languages we measure here.
\section{Hardware and software}\label{sec:technical_details}

All experiments were run on 1-2 NVIDIA GeForce RTX 3090 GPUs with 24GB of memory, using a small amount of compute and memory from an additional GPU for OSST with Mistral 24B. We used PyTorch 2.7.1 with CUDA 12.2. Code is available at \repolink, as well as the versions of the libraries as an \texttt{uv} environment.
\section{Do we need to frame as a rewriting problem instead of a continuation one?}\label{sec:no_neutral}

\begin{figure*}[!ht]
    \centering
    \includegraphics[width=0.9\linewidth]{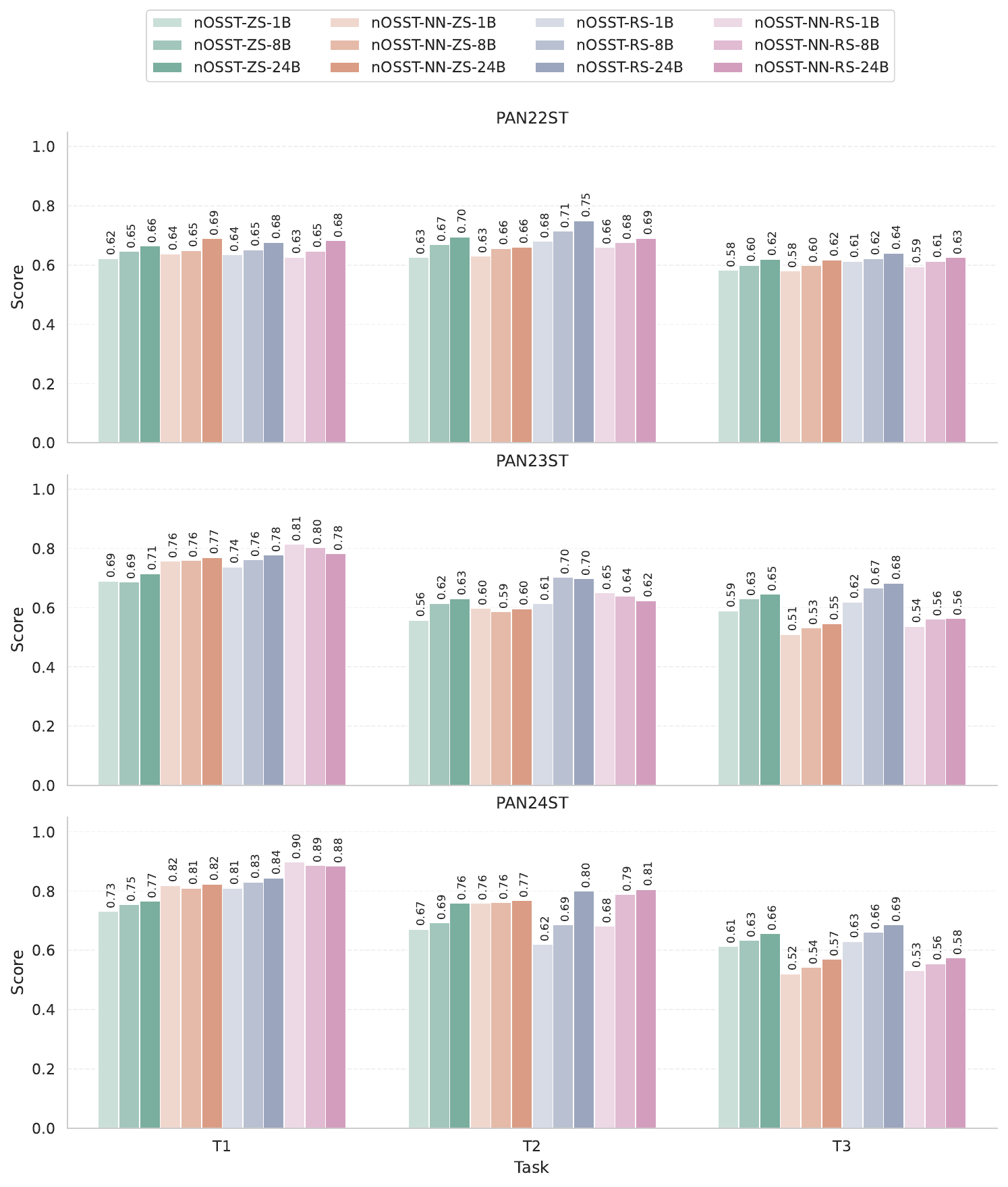}
    \caption{Comparison between using (nOSST) and not using (nOSST-NN) a neutral version to compute log-probabilities.}
    \label{fig:nn}
\end{figure*}

A natural question is whether framing the problem as rewriting is actually required. To assess this, we compare our approach against a baseline \emph{(NN)} that directly continues the input text without a neutral version to rewrite. As shown in \Cref{fig:nn}, omitting the neutral version has little impact (and can even be advantageous) on datasets where topical cues are strongly aligned. However, in settings with minimal topical correlation, performance drops sharply. This is unsurprising: generating a continuation of one text conditioned on another is substantially more difficult when their topics diverge, especially in the initial tokens, which makes distinguishing related from unrelated pairs significantly easier.

\section{To what extent does the neutralizer model influence performance?}\label{sec:neutralizer_model}

\begin{figure*}[!ht]
    \centering
    \includegraphics[width=0.9\linewidth]{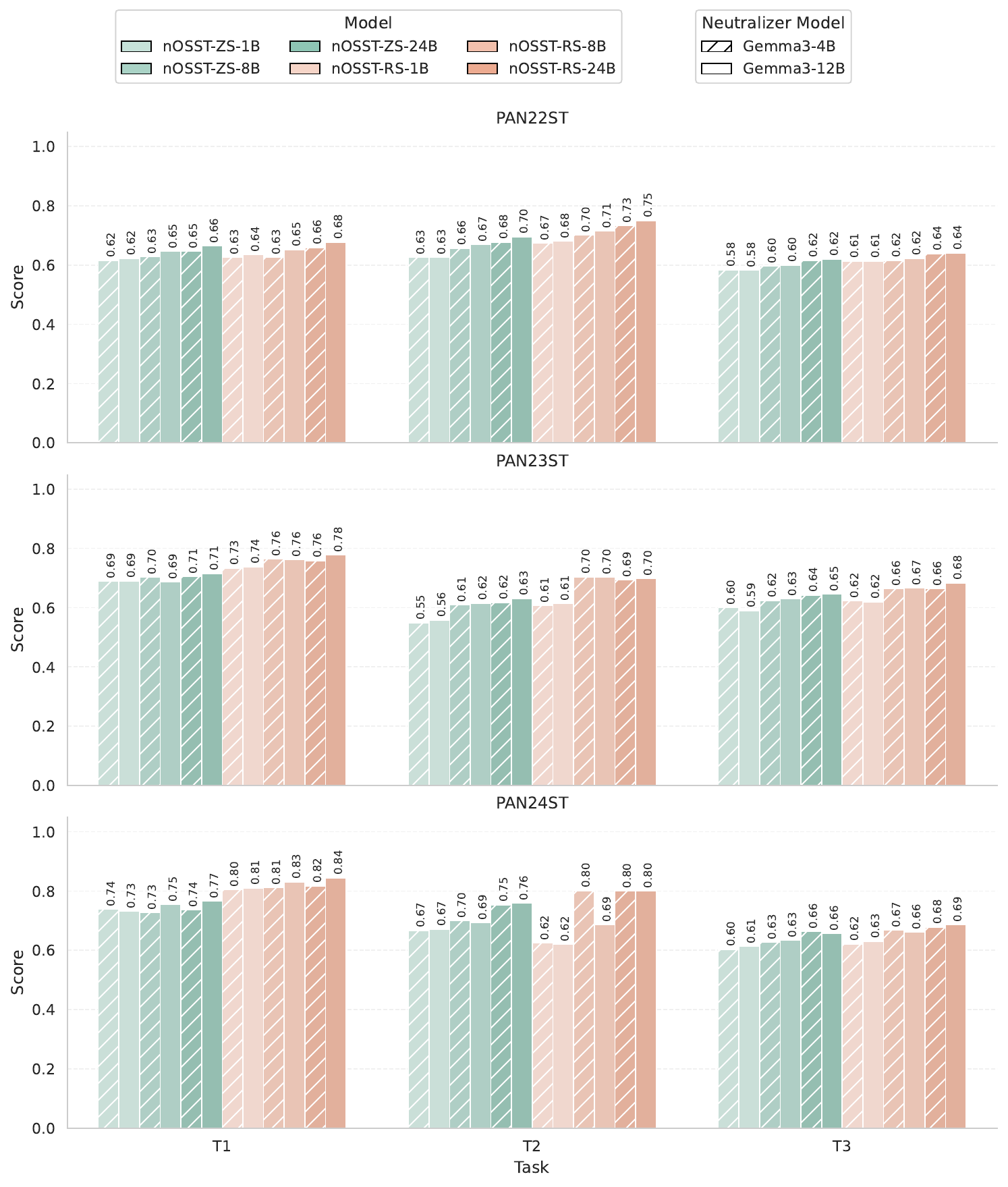}
    \caption{Comparison between using a \texttt{google/gemma-3-4b-it} \texttt{google/gemma-3-12b-it}~\cite{team2025Gemma3} model to create neutral versions of texts in PAN22ST, PAN23ST and PAN24ST. RS anchors are kept equal from the 12b neutralizer.}
    \label{fig:neutralizer-model-ablation}
\end{figure*}

In this work we have used a model of 12B parameters to create neutral versions of the given texts. This choice greatly influences the amount of resources needed to run our model overall. Is this an important choice? In \Cref{fig:neutralizer-model-ablation} we show a comparison between using a 4B and a 12B model in PAN22-24ST tasks. As we can see, this downgrade does affect performance slightly, but not to a great extent.

\section{Metrics}\label{sec:metrics}
We use the following metrics to evaluate models in this work. For closed-set authorship attribution we report the \textit{accuracy} (the ratio of times the correct candidate is picked) and the \textit{rank of the correct candidate} (with $1$ being the highest possible rank and $0$ the lowest). For open-set attribution we also include the \textit{macro F1 score} of the binary classification problem of deciding whether the true author is among the candidates. Finally, for verification we simply report the F1 score of the binary classification problem of deciding whether the two texts are written by the same author.

\end{document}